%% file: main.tex
\documentclass{article}
\usepackage{amssymb}
\usepackage{amsmath} 
\usepackage{enumitem}
\usepackage{graphicx}
\usepackage{wrapfig}
\usepackage{caption}
\usepackage{dsfont}

\usepackage[final]{corl_2025} 

\usepackage{enumitem}
\usepackage{soul}

\newlist{compactenumerate}{enumerate}{1}
\setlist[compactenumerate]{nosep, topsep=0pt, left=3pt, label=\arabic*.}

\newlist{compactitemize}{itemize}{1}
\setlist[compactitemize]{nosep, topsep=0pt, left=3pt, label=\textbullet,}

\usepackage[toc,page,header]{appendix}
\usepackage{minitoc}

\usepackage{xcolor}
\usepackage{subcaption}
\usepackage{booktabs}

\title{Learn from What We HAVE: \underline{H}istory-\underline{A}ware \underline{VE}rifier that Reasons about Past Interactions Online}

\author{
  Yishu Li$^{1}$, Xinyi Mao$^{2}$, Ying Yuan$^{1}$, Kyutae Sim$^{1}$, Ben Eisner$^{1}$, David Held$^{1}$ \\
  $^{1}$Robotics Institute, Carnegie Mellon University \\ $^{2}$Computer Science and Technology, Tsinghua University\\
}

\begin{document}
\doparttoc 
\faketableofcontents
\maketitle

\begin{abstract}
    We introduce a novel History-Aware VErifier (HAVE) to disambiguate uncertain scenarios online by leveraging past interactions. Robots frequently encounter visually ambiguous objects whose manipulation outcomes remain uncertain until physically interacted with. While generative models alone could theoretically adapt to such ambiguity, in practice they obtain suboptimal performance in ambiguous cases, even when conditioned on action history.  To address this, we propose explicitly decoupling action generation from verification: we use an unconditional diffusion-based generator to propose multiple candidate actions and employ our history-aware verifier to select the most promising action by reasoning about past interactions. Through theoretical analysis, we demonstrate that employing a verifier significantly improves expected action quality. Empirical evaluations and analysis across multiple simulated and real-world environments including articulated objects, multi-modal doors, and uneven object pick-up confirm the effectiveness of our method and improvements over baselines. Our project website is available at: \href{https://liy1shu.github.io/HAVE_CoRL25/}{https://liy1shu.github.io/HAVE\_CoRL25/}.
\end{abstract}




\section{Introduction}
\label{sec:intro}
	
In the real world, objects that appear visually similar can exhibit vastly different physical behaviors. A closed door might look identical regardless of whether it opens by pushing or pulling, and two boxes with the same outward appearances may demand different manipulation strategies depending on how their inside contents are distributed. A visual policy without history-based adaptation, 
no matter how large, would struggle 
to disambiguate such cases. 
When encountering a new object, we want a 
robot system to explore in a trial-and-error manner, 
observe the outcomes of its interactions, and adapt its strategy accordingly. 
Our goal is to enable this kind of history-aware disambiguation, where the robot continuously updates its internal model to improve future behavior.  

Ideally, we can achieve this adaptation simply with a history-conditioned generator policy. However, our experiments with a history-conditioned diffusion model led to suboptimal performance.  
Indeed, recent work has demonstrated the theoretic and empirical difficulty of training a policy (via supervised fine-tuning) under settings in which the ground-truth solution traces are heterogeneous for the same input~\cite{setlur2025scaling}. This prior work demonstrates (theoretically and empirically) the benefits of instead using verifier-based methods, even with learned rewards. 



Inspired by this work, we propose, for ambiguous object manipulation,  to train a history-aware verifier that takes in the history interactions and estimates a reward for action proposals sampled from a probabilistic action generator. We demonstrate, theoretically and empirically, the benefits of this approach compared to using only a history-conditioned generator policy.  We show theoretically that using any better-than-random-chance verifier to select an action proposed by a generator policy can improve the policy performance. We also demonstrate  the effectiveness of our approach in different simulation and real world environments with fundamental ambiguity
including opening ambiguous articulated objects, ambiguous doors, and picking up objects with an uneven mass distribution.  
We also perform ablations and analysis to better understand the importance of various design choices of 
our proposed method. 
The contributions of our paper include:

\begin{compactenumerate}[label={(\arabic*)}]
    \item A novel verifier structure that reasons about past interactions for disambiguation.
    \item Theoretical  analysis of using a verifier to select actions.
    \item Simulated and real world experiments in multiple environments with fundamental ambiguities, such as opening ambiguous articulated objects, ambiguous doors, and lifting objects with uneven mass distributions.
\end{compactenumerate}


\section{Problem Statement and Assumptions}
\label{sec:statement}


In this paper, we aim to tackle the problem of learning a policy in a Partially Observable Markov Decision Process (POMDP), in which the agent does not observe the true state but instead receives an observation 
 that provides partial information about the state. 
In such a case,  the history of observations and actions with the environment is essential for being able to estimate the true underlying state of the environment, which is important for effective decision-making. Therefore, leveraging past interactions—both successful and failed—can help an agent make more informed decisions. 

Formally, the environment is characterized by a distribution over POMDPs. More concretely, 
consider a POMDP defined over a state space $\mathcal{S}$, an action space $\mathcal{A}$, and an obervation space $\mathcal{O}$, with a transition function $\mathcal{T}: \mathcal{S} \times \mathcal{A} \to \Delta(\mathcal{S})$, 
where $\Delta(\mathcal{S})$ denotes the set of probability distributions over $\mathcal{S}$, and $\mathcal{T}(s' \mid s, a)$ denotes the probability of transitioning to state $s'$ after taking action $a$ in state $s$.
For each POMDP, assume that the (unknown) transition function $\mathcal{T}$ is sampled from a distribution $\mathcal{P}_{\mathcal{T}}$ over possible transition dynamics, that is, $\mathcal{T} \sim \mathcal{P}_{\mathcal{T}}$.
For example, consider a family of door-opening tasks, where the agent cannot determine from a single observation $o \in \mathcal{O}$ whether a door should be pushed or pulled. However, through interactions with the environment, the agent can use the history of observations and actions to implicitly infer the underlying transition dynamics of the current instance of the POMDP.
 Our task for the agent to use this history of observations and actions to discover the optimal policy for each POMDP.

\section{Related Works}
\label{sec:related-works}

\textbf{Interactive Perception}: Prior works have explored how to use interactions to actively acquire informative observations to infer about hidden physical properties or dynamics in various robotics contexts~\cite{bohg2017interactive, huang2023went} from articulation~\cite{katz2008articulate} to deformable objects~\cite{weng2024interactive}. More recently, in-context learning methods have been used to adapt robot behavior online by conditioning on recent interactions~\cite{allevato2020tunenet, xu2022prompting} through reinforcement learning~\cite{laskin2022context, wang2025adamanip} or directly modeling simulation parameters~\cite{zhang2025dynamics}. Our work builds on this idea, introducing a generator-verifier system that decomposes policy generation and verification, and explicitly evaluates action candidates using prior interactions, enabling efficient and robust online disambiguation in ambiguous, multi-modal settings.

\textbf{Generation-Verification Paradigm}: Many recent advances in machine learning build on a generation-verification paradigm, from adversarial learning in a GAN~\cite{goodfellow2020generative}, a policy generator and reward model in reinforcement learning and RLHF~\cite{christiano2017deep, bai2022training, ouyang2022training, swamy2025all}, to test-time scaling in LLMs~\cite{cobbe2021training, setlur2024rewarding, hosseini2024v, rafailov2023direct, singh2023beyond, setlur2025scaling}. Many of these work~\cite{setlur2025scaling, song2024mind, swamy2025all} stress the importance of verification model from different theoretical views. In parallel, some safety-focused approaches in robotics~\cite{ames2019control, wang2024inference, jeong2024robots, nakamura2025generalizing, wu2025foresight, borquez2025dualguard} have adopted a similar structure with a nominal generator / policy and a verifier / filter to ensure constraint satisfaction at runtime. Aligning with these prior ideas on a high level, we focus on the theoretical explanation and experimental analysis on the benefit of a system with both a generator and a verifier.


\section{Theoretical Motivation}


The main idea behind our approach is as follows:
Given an (unobserved) state space of $S$, an observation space $O$, and an action space of $A$, 
our proposed framework consists of a generative policy $\pi_G: O \times A \to [0, 1]$ that models the distribution of actions given an observation, and a verifier (or learned reward function) $V: O \times A \to \mathbb{R}$ that evaluates each of the proposed actions and outputs an estimated (scalar) reward for each one.
During inference, we sample a batch of $N$ actions $\{a_1, \ldots, a_N\}$ from the generator, $a_i \sim \pi_G(s)$. 
We predict the reward of each sampled action with the verifier $V(a_i)$ and then execute the action with the highest estimated reward. Below we present some theoretical justifications for this approach.

\subsection{Generation-Verification Gap}

Our approach is inspired by the theoretical result of \citet{setlur2025scaling}, which describes a setting in which the correct solution traces for a given input are hetereogeneous (as defined in Property 5.2 in \citet{setlur2025scaling}).  This situation is analogous to our setting in which there is a partially observed environment; 
in such a case, there will be diverse optimal actions for the same observation (in which the true optimal action depends on the unobserved ground-truth state of the environment).





\citet{setlur2025scaling} then show that, in such cases,
 training a generative policy to imitate a hetereogeneous expert dataset will induce a reward gap of $\Omega\Big(\sigma_e\sqrt{1/n}\Big)$, where $n$ is the amount of expert data, and $\sigma_e$ is variance describing the heterogeneity in the expert solution traces (Theorem 5.4). In our problem setting (Sec.~\ref{sec:statement}), the ground-truth differs given the same partial observation due to different underlying transition function, leading to a large $\sigma_e$.
On the other hand, they 
show that if we train a verifier to fit the reward terms in the expert dataset, 
we can achieve an expected error that contracts at a rate of $\widetilde{\mathcal{O}}_{H}\!\left(
     H/n
 \right)$ (Proposition 5.5), where $H$ is the horizon length.
 Therefore we can see that using a verifier is asymptotically more data-efficient as the reward gap contracts with the scale of $1/n$ while the reward gap in the generation model only contracts with a scale of $\sqrt{1/n}$.  
 





\vspace{-5pt}
\subsection{Verifier-based Selection}
\label{sec:theory_main_text}

Our manipulation framework 
uses a verifier 
to 
select the best action sampled from an action generator, compared to just sampling from the generator once. 
Below we provide a theoretical justification for this idea. 
In the below analysis, we use the simplified ``non-contextual" setting in which the generator and verifier are not conditioned on the state.
Consider a task with binary rewards. 
The generator \( \pi_G\)
represents a distribution over the action space $A \subseteq \mathbb{R}^n$ where we can sample actions from, the ground truth reward model \( R_{gt}: A \to \{0, 1\} \) maps actions \( a \in A \) to binary rewards, and the verifier \( V: A \to \{0, 1\} \) maps actions \( a \in A \) to estimated binary rewards.
In reality, both the generator  and the verifier are trained models that make errors with some probability.  
We thus make the following assumptions: 

\begin{compactenumerate}
    \item The accuracies of the generator and the verifier are independent.
    \item The generator has probability $p_G$ of sampling an action with reward 1: $P\left(R_{\text{gt}}(a) = 1 \mid a\sim \pi_G\right) = p_G$.
    \item The verifier has probability $p_V$ of predicting the correct reward for any action:  $P\left(V(a) = R_{\text{gt}}(a)\right) = p_V$, $\forall a\in A$. 
\end{compactenumerate}

We compare the expected reward of 2 methods:

\begin{compactitemize}
    \item Method 1: Sample one action from the generator: $a_{\text{naive}} \sim \pi_G$.

    \item Method 2: Sample $\mathrm{N}$ actions ${a_1, \ldots, a_\mathrm{N}}$ from the generator and use the verifier to choose the action with the largest estimated reward: $a_{\text{w/ver}}=a_{i^*}$, where $i^* = \arg \max_{i \in \{1, \ldots, \mathrm{N}\}} V(a_i)$. If multiple actions have the same highest score predicted by the verifier, randomly pick one of them.
    
    
\end{compactitemize}

For Method 1, the expected reward can be computed as: 
$\mathbb{E}\left[R_{\text{gt}}(a_{\text{naive}})\right] = p_G \times 1 + (1-p_G) \times 0 = p_G$

For Method 2,  the expected reward can be computed as: 
$\mathbb{E}\left[R_{\text{gt}}(a_{\text{w/ver}})\right] 
= \left(1-(1-Q)^\mathrm{N}\right)\frac{p_Gp_V}{Q} + (1-Q)^\mathrm{N}\frac{p_G(1-p_V)}{1-Q}$,
$\text{where } Q = P\left(V(a)=1\right) = (1-p_G)(1-p_V) + p_Gp_V.$
Then we can prove (refer to the Appendix \ref{sec:theory_discrete_independent} for full proof) that with $\mathrm{N}>1$ and $p_V > 0.5$, we have $\mathbb{E}\left[R_{\text{gt}}(a_{\text{w/ver}})\right] > \mathbb{E}\left[R_{\text{gt}}(a_{\text{naive}})\right]$. This demonstrates that our verifier is expected to be useful as long as it has an accuracy greater than 50\%. As a numerical example, if we let $\mathrm{N} = 2$, $p_G=p_V=0.9$, we have $\mathbb{E}\left[R_{\text{gt}}(a_{\text{naive}})\right] = 0.9$ and $\mathbb{E}\left[R_{\text{gt}}(a_{\text{w/ver}})\right] = 0.97$, meaning that the failure rate (the probability of executing an action with 0 reward) is reduced from 10\% to 3\% even with only one extra sample. See Appendix \ref{sec:theory_appendix} for the full proofs, a theoretical extension that removes the  independence assumption, an extension to continuous rewards, and numerical examples for both discrete and continuous cases.


\vspace{-3pt}
\section{Method}
\label{sec:method}
\vspace{-2pt}

Based on the above motivation, we construct our method, which consists of an unconditional diffusion-based generator and a history-aware verifier. We demonstrate in our experiments that such an architecture leads to much better performance than a history-conditioned diffusion model without a verifier, as also suggested by the above theoretical analysis.

Given the action generator \( G_{\phi} \), at each timestep \( t \), we sample action proposals \( a_t^{(0)}, \dots, a_t^{(\textrm{N}-1)} \) based on the current observation \( o_t \). The history-aware verifier \( V_{\theta} \) then outputs corresponding scores \( s_t^{(0)}, \dots, s_t^{(\textrm{N}-1)} \) for the proposed actions, considering the history of past actions and their outcomes, \( (o_0, a_0, \dots, o_{t-1}, a_{t-1}, o_t) \). Finally, we select and execute the action with the highest estimated score \( a_t^{(\hat{i})} \), where \( \hat{i} = \arg\max_i s_t^{(i)} \). Below we describe the details of our approach.

\vspace{-5pt}
\subsection{Dataset Generation}

We desire for our method to be able to quickly adapt its future actions after performing either high-reward or low-reward actions and observing the results. 
Thus, we generate an offline dataset for training where each sample consists of a history sequence and an action proposal set. We generate each history sequence by either sampling a random action (most likely leading to a failure) or using the ground truth action (obtained using privileged knowledge in the simulator) at each step. 
The action proposal set is constructed with the ground-truth action, random actions and history actions. We label the score for each action proposal and history actions; 
 see Appendix \ref{sec:dataset_details} for details.


\subsection{History-Aware Verifier}
\label{sec:History-Aware Verifier}


We assume that we are given an action proposal $\hat{a}$ as well as  history actions and observations.  Specifically, to learn from history, we group the data based on the observation before an action and the resulting observation after an action: 
$\{(o_{i-1}, a_i, o_i), i \in [t-1]\}$.  We  compute action embeddings $E_{\textrm{action}}: A\to \mathbb{R}^n$ (where $n$ is the encoded action's feature dimension) to obtain an embedding $\hat{a}_{\textrm{emb}} = E_{\textrm{action}}(\hat{a})$ for the action proposal $\hat{a}$ as well as an embedding for each action in the action history $\{a_{\textrm{emb},i} = E_{\textrm{action}}(a_i), i\in[t-1]\}$.
We also encode each pair of subsequent observations  $E_{obs}: O \times O \to \mathbb{R}^m$ (where $m$ is the observation feature dimension) to obtain observation embeddings: 
$\{f_i = E_{\textrm{obs}}(o_{i-1}, o_i), i\in[t-1]\}$. 
We then input the action embeddings for the proposal as well as the history through an action transformer to receive an updated action proposal embedding $\hat{a}_{\textrm{emb}}'$ and new history action embeddings 
$a'_{\textrm{emb},i}$.
We also input the observation embeddings into an observation transformer to receive updated observation history embeddings $f'_i$. The action proposal embedding $\hat{a}_{\textrm{emb}}$ is also passed through an unconditional action encoder to obtain an unconditional evaluation of the proposed action without conditioning on history: $\hat{v}_u = E_{\textrm{uncond}}(\hat{a}_{\textrm{emb}})$ (used below).

Given a history of length $t$, we perform an explicit dot-product attention operation with $\hat{a}_{\textrm{emb}}'$ as the query ($Q$), action history embeddings $\{a'_{\textrm{emb},i}\}_{i=1}^t$ as keys ($K$) and the observation history embeddings $\{f'_i\}_{i=1}^t$ as values ($V$): $\hat{v} 
= \text{softmax}\left( \left[QK^\top, \mu_{qk}\right] / \sqrt{d_k} \right) [V, \hat{v}_u]$,
where $d_k$ is the dimension of key embeddings, and $[\cdot, \cdot]$ denotes concatenation along the key/value dimension. Here $\mu_{qk}$ denotes a constant  to represent the logit associated with the unconditional evaluation of the proposed action $\hat{a}$ (without regard to the history), which we compute as a running mean of the $QK^T$ logits during training. 
The intuition behind this explicit attention layer is that we want the model to relate the action proposal $\hat{a}$ (encoded as the query  $Q = \hat{a}_{\textrm{emb}}'$) with similar actions in the history (encoded as the keys $K = \{a'_{\textrm{emb},i}\}$) and refer to their action results (encoded as values $V = \{f'_i\}$).  The unconditional logit $\mu_{qk}$ and embedding $\hat{v}_u$ exists to handle cases where there are no  actions similar to the proposal $\hat{a}$ in the history (or no actions in the history at all). 
Then the final embedding $\hat{v}$ will go through a score decoder (small MLP with tanh on the output) to obtain a 1-dim logit as the final output score: $s = D_{\textrm{score}}(\hat{v})\in [-1, 1]$. We compare our proposed architecture to a monolithic transformer in Sec.~\ref{sec:Analysis} and show improved performance with our approach.

\begin{figure}[h]
    \centering
    \includegraphics[width=0.9\textwidth]{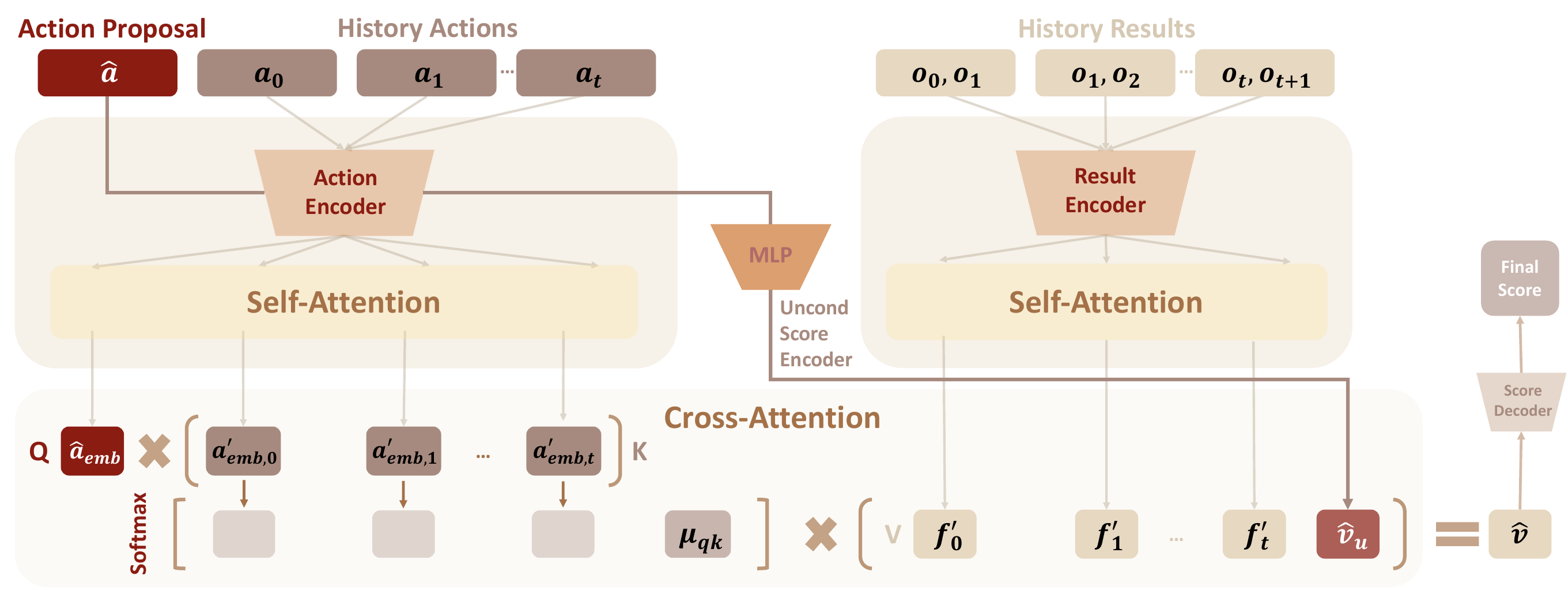} 
    \caption{\textbf{HAVE Architecture}: the proposed and history actions are encoded through a PointNet++ for 3D geometric understanding, and self-attention layers for sequential reasoning. History results are encoded similarly. Together they pass through an explicit attention layer to obtain the final score.}
    \label{fig:model}
\end{figure}

During training, we supervise the final predicted score $s$ for the action proposal, using an MSE loss with the ground-truth scores (referred to as ``Final Score Loss" in Fig.~\ref{fig:model}).  We also supervise the unconditional embedding $\hat{v}_u$ and the action results history embeddings $\{f'_i\}_{i=1}^t$ by passing them through the  score decoder $D_{\textrm{score}}$ and supervising them with ground truth rewards for corresponding timestep (referred to as ``Unconditional Score Loss" and ``History Score Loss" respectively in Fig.~\ref{fig:model}).

\subsection{Network Architecture}
\label{sec:Network Architecture}

We represent actions as dense action fields, where given a point cloud $\{p_i\}$, and an action point $p$ and direction $d$, the dense action field is the point cloud with 3 extra channels to represent the action, calculated as $\{d_i: d * \exp\{-c * ||p_i - p||_2^2\}$ where $c$ is a scaling constant. This representation also makes actions state-aware, enabling reasonable understanding about relationships between actions executed in different states. We hypothesize that this dense action representation is easier to process than a sparse action representation, as indicated by our ablation experiments.  We then encode this dense action representation $a_{\textrm{emb},i} = E_{\textrm{action}}(a_i)$ using a PointNet++~\cite{qi2017pn2} point cloud encoder.

\vspace{-5pt}

We represent the observation encoder $E_{\textrm{obs}}(o_{i-1}, o_i)$ as follows: First, we compute the flow from each point in observation $o_{i-1}$ to the corresponding location of that point in observation $o_i$, which we refer to as the ``Observation Flow."  In  training we use the ground-truth flow, and at test time  we use a state-of-the-art 3D point tracker~\cite{ngo2024delta}. We then create a point cloud with 3 extra channels for the 3D flow, which we also encode with a PointNet++~\cite{qi2017pn2} point cloud encoder.  We construct the action generator 
similar to prior work~\cite{li2024flowbothd}, but without history injection; see Appendix \ref{sec:model-details} for details.



\section{Experiments and Analysis}
\label{sec:experiments}

To evaluate our system's capability to solve ambiguous robotics tasks, we carried out experiments in 3 different environments: opening general articulated objects, opening ambiguous doors, and lifting objects with uneven mass distributions. We perform experiments both in simulation environments and in a real-world robot setup. At inference time, the robot incrementally builds its history through each interaction with the object.

\subsection{Articulated Objects}

\textbf{Baselines}: We compare our method to the following baselines: (1) FlowBot3D~\cite{eisner2022flowbot3d}, (2) FlowBotHD~\cite{li2024flowbothd} with and without ``CC" (the heuristic consistency check in FlowBotHD that filters the proposed actions), (3) Generator Only: unconditional diffusion without a verifier, (4) Conditional Generator: conditional diffusion that takes in the same history interaction information as the verifier, we use classifier-free guidance~\cite{ho2022classifier} for conditioning (see Appendix \ref{sec:cond_diff} for architecture details, and \ref{sec:cfg_for_cond_diff} for different cfg scales). For all baselines and our own policy, we train the action generator  with a training dataset augmented with fully closed object examples~\cite{li2024flowbothd}, and adopt the FlowBotHD switch grasp policy  to enable models to adapt from wrong grasp points (see \citet{li2024flowbothd} for details). 

\textbf{Ablations}: We also perform the following ablations to better understand the effect of our design choices: (1) \textit{w/o History Score Loss}: Train without  the ``History Score Loss" (see Sec.~\ref{sec:History-Aware Verifier} and Figure~\ref{fig:model}); (2) \textit{w/o Uncond Score Loss}: Train without  the ``Unconditional Score Loss" (see Sec.~\ref{sec:History-Aware Verifier} and Figure~\ref{fig:model}); (3) \textit{Sparse Action}: rather than representing an action as a dense action field (Sec.~\ref{sec:Network Architecture}), we represent the action as a point cloud with only one action vector at the contact point; (4) \textit{Point Cloud as Result}: directly use point cloud sequence as action result without explicit extracting observation flow (Sec.~\ref{sec:Network Architecture}). Please refer to Appendix \ref{sec:pcd_as_result} for more details.
As stated in Sec.~\ref{sec:Network Architecture}, for the observation encoder $E_{\textrm{obs}}(o_{i-1}, o_i)$ we compute the flow from each point in observation $o_{i-1}$ to the corresponding location of that point in observation $o_i$.  
The ablations all use the ground-truth flow at test time (obtained from the simulator), to give the ablations the best chance to succeed. For a fair comparison, we show the performance of our method both using ground-truth flow (``HAVE (Ours) + GT obs flow") as well as our method using estimated flow (``HAVE (Ours) + Estimated flow") using the 3D tracker DELTA~\cite{ngo2024delta}.  

\subsubsection{General Articulated Objects in Simulation}

We follow \citet{li2024flowbothd} and split PartNet-Mobility into train and test instances, and construct a simulation environment with a suction gripper. We initialize each object at a fully closed state, and the task of the policy is to fully open the objects. In Table \ref{tab:failure-all}, we show the \textbf{Failure Rate}: the percentage of objects that are opened less than 90\% within 30 steps. Given the stochasticity  in the generative model, we test each  object 5 times. We also report experiments with the split following \citet{eisner2022flowbot3d} where we hold-out 10 categories during training and test on them (Appendix \ref{sec: heldout_results}) .

Comparing the failure rate from our policy and the baselines, we can see that HAVE (Ours) reduces the failure rate by about 6x (from 13\% to 2\%) compared to FlowBot3D~\cite{eisner2022flowbot3d}, about 10x (from 20\% to 2\%) compared to using a history-conditioned generative model without a verifier, and about 3x (from 6\% to 2\%) compared to FlowBotHD~\cite{li2024flowbothd} even with the heurisitc consistency check (CC) policy. 
This 
validates the idea that using a generator to mimic a heterogeneous dataset is not sufficient for optimal performance without a verifier.

\textbf{Ablation Analysis:} Our ablation analysis reveals the benefit of each  of our design choices. Using a sparse action representation instead of a dense action field (\textit{Sparse Action}) decreases the performance most. 
Using point cloud sequences without extracting observation flows (\textit{Point Cloud as Result}) also underperforms using explicit observation flow. Training without the ``History Score Loss" or the ``Unconditional Score Loss"  leads to a slight performance drop, indicating that supervising the history observation and the unconditional action embedding benefits the model.

\input{tables/fr_train-val_new}

\subsubsection{Multimodal Doors in Simulation}

\begin{figure}[h]
    \centering
    \includegraphics[width=1.0\textwidth]{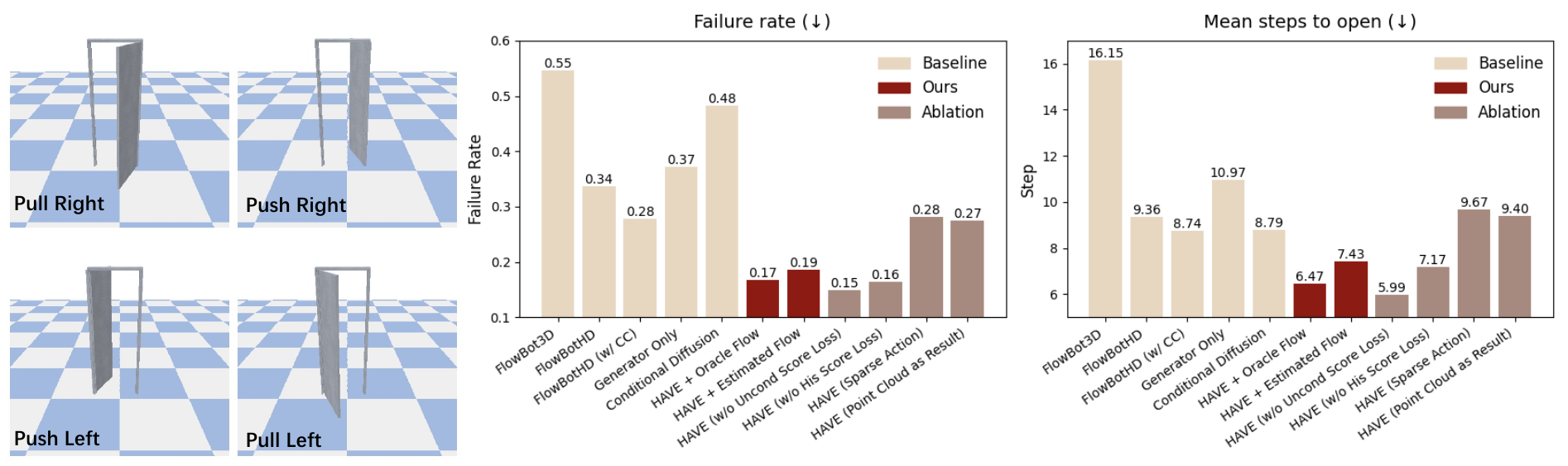} 
    \caption{\textbf{Multi-modal Door Dataset Performance}: The left plot shows an example of a constructed multi-modal door in simulation with the same geometry but different opening directions. The right bar plot demonstrates the efficacy metric \textbf{Failure Rate} and the efficiency metric \textbf{Steps to Open} from which we can see our method's improvements over baselines and ablation architectures.}
    \label{fig:door_sim}
\end{figure}

To further test our method's performance in an ambiguous setting, we also create a multimodal dataset in simulation with doors from PartNet-Mobility. We create multiple duplicates of each door with the same geometry but different joint positions and rotation directions (corresponding to 4 different opening directions - push left, push right, pull left, and pull right). For doors with handles, we create 2 duplicates with handles (varying whether the door is push or pull), and we create 4 duplicates with the handle removed. In Figure~\ref{fig:door_sim}, we show an example of a multi-modal door in simulation. 
In addition to \textbf{Failure Rate} we also compute the \textbf{Mean steps to Open} (the steps taken to open the door to 5\%) as a metric of efficiency of disambiguating different modes with exploration. From Figure~\ref{fig:door_sim}, we can see that our method significantly improves the performance (reduces the failure rate and step number) on ambiguous doors compared to prior approaches and ablations (except for the minor design variant ``w/o Unconditional Score Loss"), further validating our approach.  


\subsubsection{Real World Ambiguous Door}

\begin{figure}[h]
    \centering
    \includegraphics[width=1.0\textwidth]{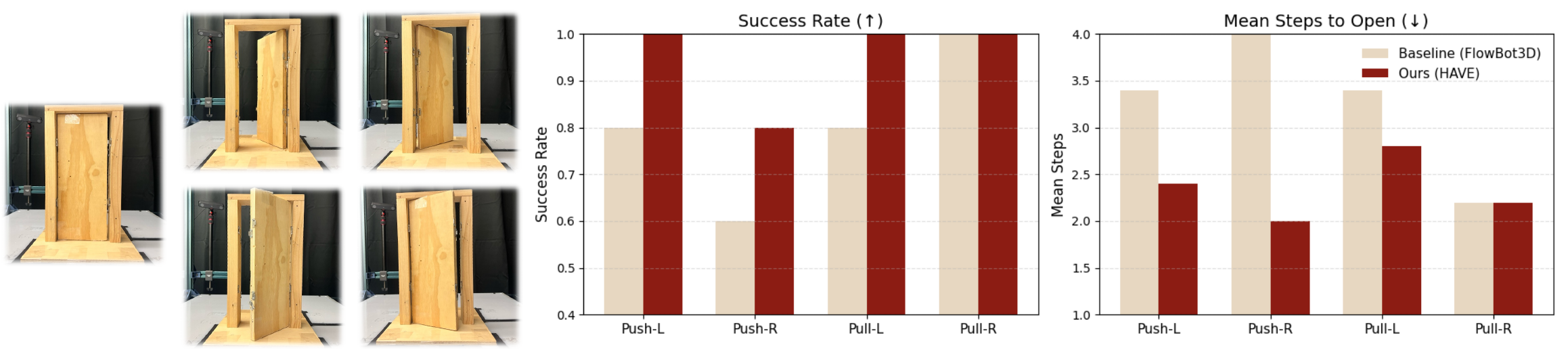} 
    \caption{\textbf{Real World Ambiguous Door Performance}:  The left side is the visual appearance of the ambiguous door and the 4 modes it opens. We collect 5 trials for each mode. From the bar plot we can see that HAVE demonstrates a more stable and efficient opening process.}
    \label{fig:door_real_world}
\end{figure}

We apply our method in the real world on a custom-made multi-modal door, similar to the multimodal door used in \citet{li2024flowbothd}.  To manipulate this object, we use a Franka Emika Panda Robot with a Schmalz Cobot Pump as an end effector.  
We give each policy a maximum of 5 steps to succeed; see Appendix \ref{sec:real_world_details} for further details.  
We can see from Figure~\ref{fig:door_real_world} that HAVE (Ours) has a greater success rate over the different modes with a lower number of  mean steps to open.

\subsection{Uneven Object Pick-up}

The task is to lift objects with unevenly distributed mass (unknown mass center). We constructed a simulation environment with a Franka gripper equipped with a spatula, along with an ambiguous rod dataset (one rod, but different mass centers) and an unseen object dataset (unseen sizes of rods, and diverse objects), see Appendix \ref{sec:uneven_object_details} for details. 
In both cases, positions for the centers of mass are uniformly sampled along the object's length. We train on the ambiguous rod dataset and evaluate on both datasets to demonstrate the model's capability and generalization ability.

\vspace{-10pt}
\input{tables/uneven-object-val}

In Table~\ref{fig:uneven-object-table}, we show the \textbf{Failure Rate \%}: the percentage of objects failed to be lifted within 5 steps, defined as those with a tilt angle exceeding 0.01 radians or an increase in the vertical distance of the center of mass less than 0.01 meters. 
Each object is tested 5 times in the ambiguous rod dataset and 3 times in the unseen dataset. At each step, the verifier is given 20 action proposals.
From Table~\ref{fig:uneven-object-table}, we can see that on both the ambiguous rod and the unseen dataset, adding HAVE (Ours) as a verifier decreases the failure rate compared to only using either unconditional generator or conditional generator and other three baselines. 
Unlike in more complex environments like articulated objects, HAVE performs well with both generator types, likely due to a simpler action space. 
Figure~\ref{fig:uneven-object-fig} shows an example visualization of the ``Conditional Diffusion" model (left) compared to our model's output (right); 
we show the theoretical upper and lower bounds of the rod's center of mass after each attempt (the tilting direction at each step indicates whether the selected point is on the left / right side of the mass center) as bars, and the predicted positions by the model as scatter points. 



\subsection{Analysis}
\label{sec:Analysis}



\begin{wrapfigure}{r}{0.4\textwidth}   
  \vspace{-10pt}                         
  \centering
  \includegraphics[width=\linewidth]{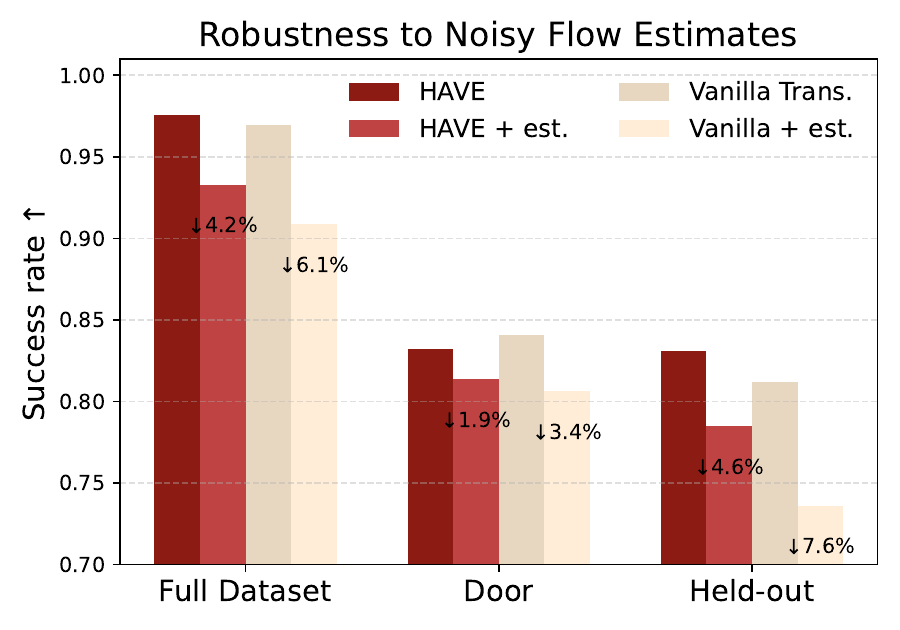}
  \caption{Comparison between \textbf{Vanilla Transformer} and \textbf{HAVE}: Robustness to noisy observation flow. }
  \label{fig:noise_robustness}
\end{wrapfigure}

\paragraph{Architecture.}
We compare our architecture with a vanilla transformer baseline where the action tokens ($\hat{a}_{emb}$, $\{a_{emb, i}\}_{i=1}^t$) and observation tokens ($\{f_i\}_{i=1}^t$) are concatenated correspondingly and processed by self-attention layers (see Appendix \ref{sec:vanilla_transformer_arch}). Specifically, we demonstrate each model's robustness to noisy estimation in the Observation Flow (Sec.~\ref{sec:Network Architecture}) in all three articulated object environment (Fullset, Multi-modal door, and Held-out). Figure~\ref{fig:noise_robustness} shows that both models perform similarly using the ground-truth flow; however,
using estimated flows (from the DELTA 3D tracker~\cite{ngo2024delta}) hurts the performance of the vanilla transformer more than our architecture (HAVE). This validates the benefit of our architectural design. 
Our intuition is that our design forces the model  to learn the relationship between query and history actions, making our model more robust to noise. 

\begin{figure}[h]
    \centering
    \includegraphics[width=1.0\textwidth]{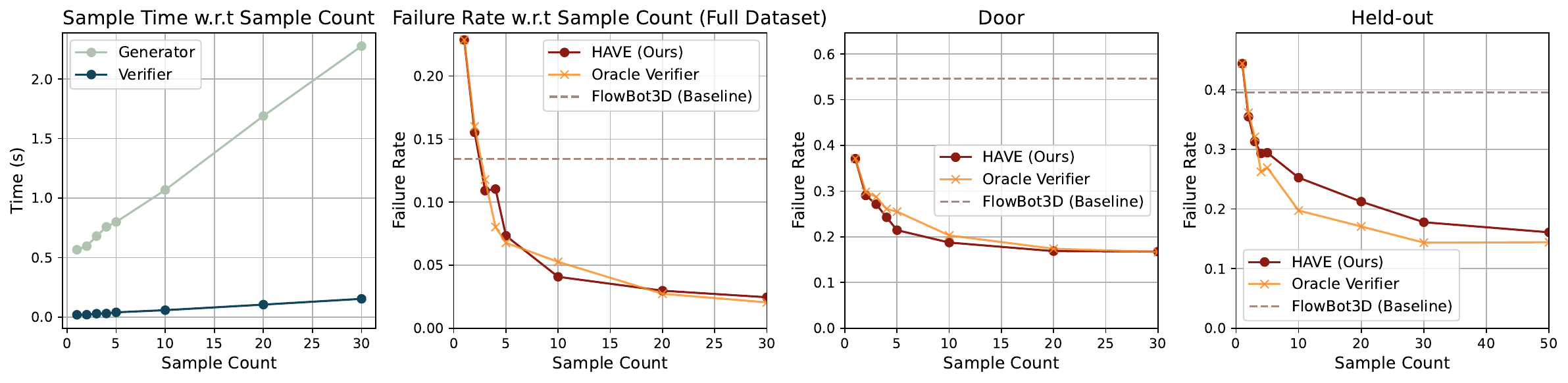} 
    \caption{\textbf{Time and Accuracy vs Number of Samples}: On the left, we plot the time used for generator and verifier w.r.t the number of generated samples (with verifier history length = 5); On the right, we plot the  failure rate w.r.t sample count, comparing HAVE with an oracle verifier (which always selects the best sample from a given batch). See Appendix ~\ref{sec:oracle_details} for details for oracle experiments.}
    \label{fig:analysis_sample_cnt}
\end{figure}

\paragraph{Number of samples:} We analyze the sample efficiency of our verifier by conducting ablations on how many samples we generate for the verifier to choose from for each step. From Figure~\ref{fig:analysis_sample_cnt}, we can see that by sampling 5 samples, the performance is already greatly improved, which demonstrates the efficiency of incorporating a verifier. 
We also plot the time cost measured with 1 RTX4090 GPU on the left to demonstrate the time required to generate and evaluate multiple samples.

\section{Conclusion}
\label{sec:conclusion}

In this work, we present HAVE, a novel history-aware verifier that reasons over past interactions to guide action selection in ambiguous manipulation tasks. By 
leveraging a verifier to select among sampled proposals, our method significantly improves efficacy and efficiency compared to baseline approaches. We provide both theoretical analysis and empirical validation across a range of simulated and real-world environments, demonstrating the benefits of our approach in handling fundamental ambiguities in manipulation. We hope that our findings can help inspire more adaptive, history-aware manipulation paradigms through combining generation and verification.

\clearpage


\section{Limitations}
\label{sec:limitation}

While our history-aware verifier demonstrated great improvements over baselines in various settings, the current generation-verification pipeline's performance is still strictly bounded by the expressiveness of the generator: if none of the generator's proposals are good, the verifier won't be able to make improvements. Furthermore, our training data generation relies on per-task defined score and privileged knowledge about the ground-truth action, typically obtained from simulation. This level of supervision may not be practically available for a diverse range of objects in real-world scenarios. In future work, we hope to further explore how the verifier can be used to actively improve the quality of the generator's generation, and how to learn score without privileged supervision.

\acknowledgments{This material is based upon work supported by the National Science Foundation under NSF CAREER Grant No. IIS-2046491 and ONR MURI N00014-24-1-2748. We thank Amrith Setlur, Aviral Kumar, and Gokul Swamy for helpful insights and discussions about generator and verifier gap, Yufei Wang for helpful discussions, Alexis Hao and Mino Nakura Fang for help with real robot experiments, and Xilun Zhang for discussion on uneven object environment setups.}


\bibliography{main}  

\input{appendix}


\end{document}

%% file: tables/fr_train-val_new.tex
\begin{table*}[ht]
\newcommand{\clipart}[1]{\includegraphics[height=6mm]{tables/icons/#1.png}}
\renewcommand{\arraystretch}{1.2}
\resizebox{\textwidth}{!}{
\setlength\tabcolsep{.2em}

\begin{tabular}{|r|c|c|cccccccccccccccccccc|}
\hline
\rule{0pt}{2.5em}
& \textbf{\underline{$\textbf{AVG}_{c}$}} & \textbf{\underline{$\textbf{AVG}_{s}$}} 
& \clipart{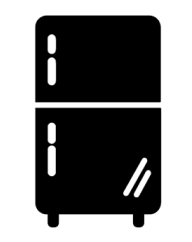} & \clipart{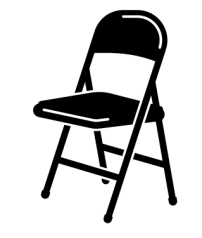} & \clipart{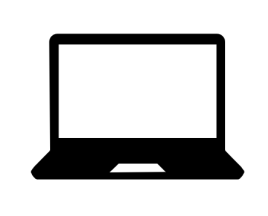} & \clipart{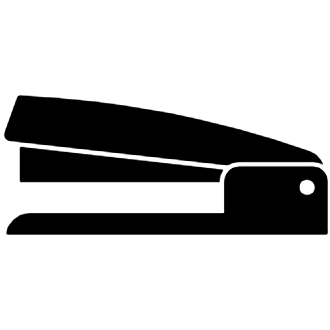} & \clipart{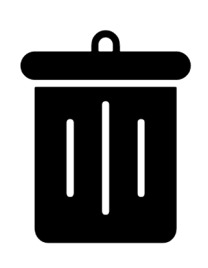} & \clipart{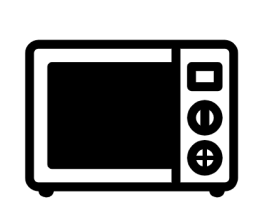} & \clipart{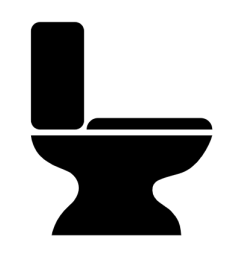} & \clipart{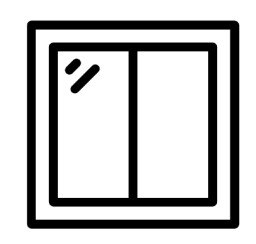} & \clipart{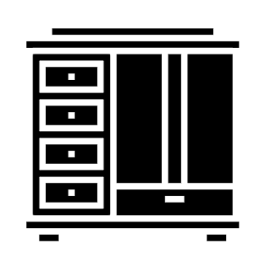} & \clipart{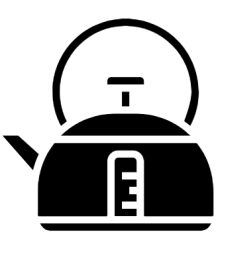}
& \clipart{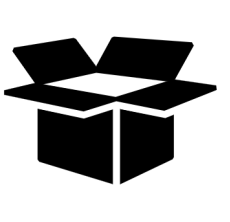} & \clipart{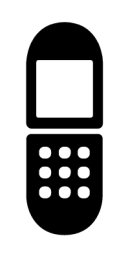} & \clipart{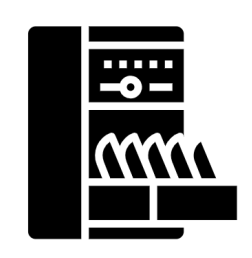} & \clipart{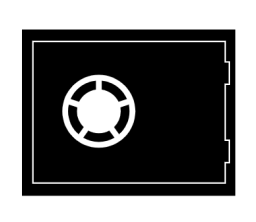} & \clipart{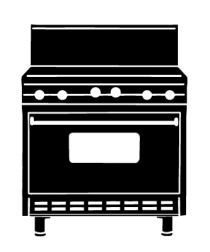} & \clipart{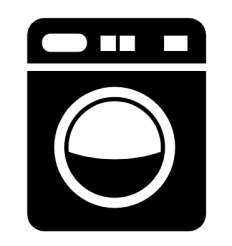} & \clipart{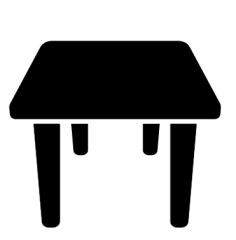} & \clipart{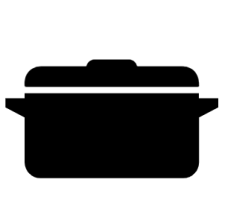} & \clipart{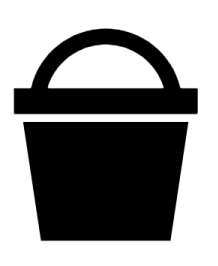} & \clipart{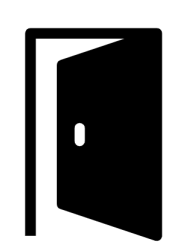} \\
\hline
\textbf{Baselines} & \multicolumn{22}{c|}{} \\
\hline
FlowBot3D & 13.4 & 5.7 & 5.3 & \textbf{0.0} & \textbf{0.0} & 4.0 & 4.0 & \textbf{0.0} & \textbf{0.0} & \textbf{0.0} & 3.8 & 20.0 & 45.0 & 100.0 & 10.0 & \textbf{0.0} & \textbf{0.0} & \textbf{0.0} & 5.3 & \textbf{0.0} & 20.0 & 51.4 \\
FlowBotHD (w/o CC) & 21.9 & 11.1 & 21.3 & 5.0 & 2.5 & 28.0 & 14.0 & 20.0 & 8.0 & \textbf{0.0} & 6.2 & 20.0 & 20.0 & 20.0 & 38.0 & 33.3 & 48.6 & 10.0 & 6.1 & \textbf{0.0} & 80.0 & 57.1 \\
FlowBotHD (w/ CC) & 6.2 & 2.9 & 1.3 & 5.0 & \textbf{0.0} & \textbf{0.0} & \textbf{0.0} & 10.0 & \textbf{0.0} & \textbf{0.0} & \textbf{0.5} & 20.0 & \textbf{0.0} & \textbf{0.0} & 20.0 & 6.7 & 5.7 & 10.0 & \textbf{2.0} & \textbf{0.0} & \textbf{0.0} & 42.9 \\
Generator Only & 22.8 & 13.6 & 49.3 & 15.0 & \textbf{0.0} & 16.0 & 4.0 & \textbf{0.0} & 8.0 & 2.0 & 10.2 & 20.0 & 25.0 & 100.0 & 16.0 & 50.0 & 2.9 & \textbf{0.0} & 9.0 & \textbf{0.0} & 90.0 & 40.0 \\
Conditional Diffusion & 19.7 & 12.6 & 24.0 & 10.0 & \textbf{0.0} & 4.0 & 8.0 & 40.0 & 4.0 & 16.0 & 9.4 & 20.0 & 60.0 & \textbf{0.0} & 18.0 & 20.0 & 34.3 & 15.0 & 7.4 & \textbf{0.0} & 30.0 & 74.3 \\
\hline
\textbf{Ours} & \multicolumn{22}{c|}{} \\
\hline
HAVE (Ours) + GT obs flow & \textbf{2.5} & \textbf{2.3} & \textbf{0.0} & \textbf{0.0} & \textbf{0.0} & 4.0 & \textbf{0.0} & \textbf{0.0} & \textbf{0.0} & \textbf{0.0} & 2.2 & 20.0 & \textbf{0.0} & \textbf{0.0} & \textbf{0.0} & 3.3 & \textbf{0.0} & \textbf{0.0} & 2.5 & \textbf{0.0} & \textbf{0.0} & 17.1 \\
HAVE (Ours) + Estimated obs flow & 6.7 & 5.6 & 13.3 & \textbf{0.0} & \textbf{0.0} & 4.0 & \textbf{0.0} & \textbf{0.0} & 2.0 & \textbf{0.0} & 5.4 & 20.0 & 25.0 & 20.0 & \textbf{0.0} & 3.3 & \textbf{0.0} & \textbf{0.0} & 5.3 & \textbf{0.0} & 10.0 & 25.7 \\
\hline
\textbf{Ours (Oracle)} & \multicolumn{22}{c|}{} \\
\hline
HAVE (Ours) + Oracle Sampler & 1.7 & 1.1 & 0.0 & 0.0 & 0.0 & 0.0 & 0.0 & 0.0 & 0.0 & 0.0 & 0.5 & 20.0 & 5.0 & 0.0 & 0.0 & 0.0 & 0.0 & 0.0 & 2.0 & 0.0 & 0.0 & 5.7 \\
\hline
\textbf{Ablations} & \multicolumn{22}{c|}{} \\
\hline
w/o Unconditional Score Loss & 3.2 & 2.9 & \textbf{0.0} & \textbf{0.0} & \textbf{0.0} & 4.0 & \textbf{0.0} & \textbf{0.0} & \textbf{0.0} & \textbf{0.0} & 2.6 & 20.0 & 10.0 & \textbf{0.0} & \textbf{0.0} & 6.7 & \textbf{0.0} & \textbf{0.0} & 4.1 & \textbf{0.0} & 10.0 & 17.1 \\
w/o History Score Loss & 3.0 & 3.0 & \textbf{0.0} & \textbf{0.0} & \textbf{0.0} & 8.0 & \textbf{0.0} & \textbf{0.0} & \textbf{0.0} & \textbf{0.0} & 3.0 & 20.0 & \textbf{0.0} & \textbf{0.0} & \textbf{0.0} & 3.3 & \textbf{0.0} & \textbf{0.0} & 4.5 & \textbf{0.0} & 10.0 & \textbf{11.4} \\
Point Cloud as Result & 8.3 & 4.1 & \textbf{0.0} & \textbf{0.0} & \textbf{0.0} & \textbf{0.0} & 8.0 & \textbf{0.0} & \textbf{0.0} & \textbf{0.0} & 1.3 & 20.0 & 50.0 & \textbf{0.0} & 10.0 & 10.0 & 5.7 & \textbf{0.0} & 3.7 & \textbf{0.0} & \textbf{0.0} & 57.1 \\
Sparse Action & 9.1 & 3.7 & \textbf{0.0} & \textbf{0.0} & \textbf{0.0} & \textbf{0.0} & \textbf{0.0} & \textbf{0.0} & \textbf{0.0} & \textbf{0.0} & 3.0 & 20.0 & 10.0 & 100.0 & \textbf{0.0} & 3.3 & \textbf{0.0} & \textbf{0.0} & 5.3 & \textbf{0.0} & 20.0 & 20.0 \\
\hline
\end{tabular}
}
\smallskip
\caption{\textbf{Failure Rate \% ($\downarrow$)}: Lower is better. AVG$_c$ = average over categories, AVG$_s$ = average over all samples. All 22 categories are treated as train categories, and we test on unseen instances. We run each object for 30 steps in simulation, and 30 samples are given for the verifier to choose from at each step. We see HAVE's overall performance increase compared with baselines and ablation choices, as well as a more balanced ability across categories. Icon-to-text correspondence in Fig.~\ref{fig:articulated_table_icons}.} 
\label{tab:failure-all}
\vspace*{-10pt}
\end{table*}

%% file: tables/uneven-object-val.tex
\begin{figure*}[ht]
    \centering
    \begin{minipage}[c]{0.3\textwidth}
        \centering
        \vspace{10pt}
        \resizebox{\textwidth}{!}{

        \begin{tabular}{|r|c|c|}
            \hline
            & \textbf{Rod} & \textbf{Unseen} \\
            \hline
            \textbf{Baselines} & \multicolumn{2}{c|}{} \\
            \hline
            FlowBot3D & 41.2 & 36.8 \\
            FlowBotHD (w/o CC) & 36.5 & 37.6 \\
            Generator Only & 36.7 & 36.8 \\
            Conditional Diffusion & 30.6 & 31.9 \\
            \hline
            HAVE + Uncond (Ours) & 32.9 & 34.5 \\
            HAVE + Cond (Ours) & \textbf{29.4} & \textbf{28.1} \\
            \hline
            \shortstack{Oracle Sampler} & 4.8 & 24.5 \\
            \hline
        \end{tabular}
        }
        \captionof{table}{Failure Rate \% given maximum steps on \textbf{Ambiguous Rod dataset} and \textbf{Unseen dataset} ($\downarrow$): Lower is better. }
        \label{fig:uneven-object-table}
    \end{minipage}
    \hfill
    \begin{minipage}[c]{0.68\textwidth}
        \centering
        \includegraphics[width=\textwidth]{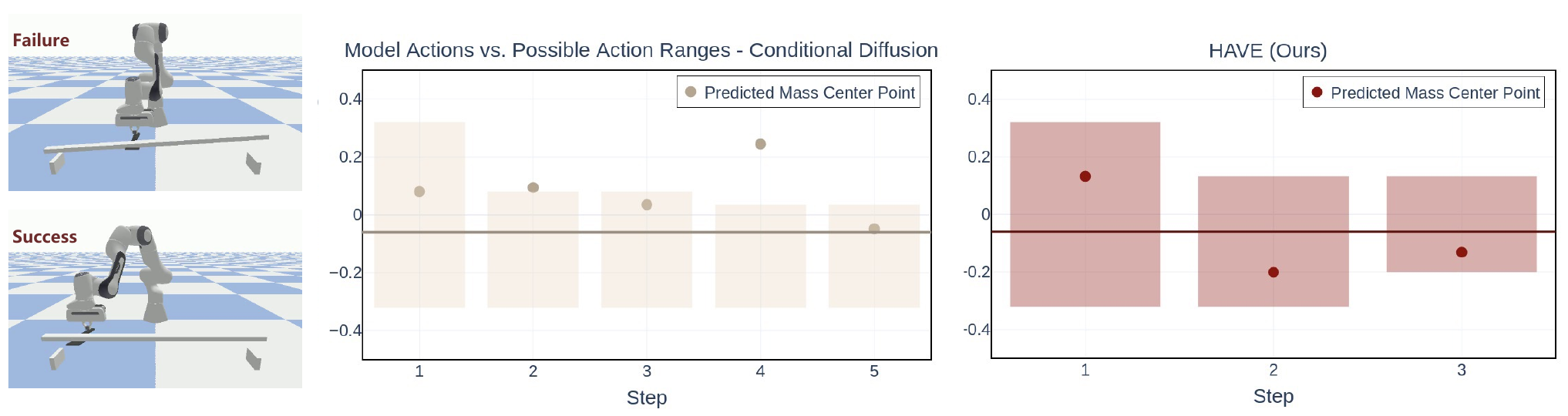}
        \caption{\textbf{Visualization of action sequence (dots)   and theoretical  center of mass range (bars)}: ``Conditional Diffusion" takes 5 steps to succeed, 
        while HAVE (Ours) takes only 3 steps and each attempt is within the theoretical center of mass range. }
        \label{fig:uneven-object-fig}
    \end{minipage}
\end{figure*}



%% file: appendix.tex
\clearpage
\newpage
\appendix
\addcontentsline{toc}{section}{Appendix} 
\part{Appendix} 
\parttoc 

\section{Theoretical Motivation}
\label{sec:theory_appendix}

\subsection{Discrete Reward with Independent Generator Verifier Accuracies}
\label{sec:theory_discrete_independent}

Let \( a \sim \pi_G \) be an action sampled from the generator, $p_G$ be the probability of the generated action having reward 1.  
Let \( R_{\text{gt}}(a) \in \{0, 1\} \) be the ground-truth reward (i.e., whether the action is good or bad), and \( V(a) \in \{0, 1\} \) be the verifier output.

We define: 
\begin{alignat*}{2}
&p_G = P\bigl(R_{\text{gt}}(a)=1 \,\bigm|\, a \sim \pi_G\bigr)\\
&p_{V} = P\left(V(a) = R_{\text{gt}}(a)\right)
\end{alignat*}
that is, \( p_{G} \) is the probability the generator generates a good action, and \( p_{V} \) is the probability the verifier predicts the correct reward given an action, \textbf{independent of the ground truth action reward} (We will show extension without this independence assumption in Theory A').

We assume that $p_G\in(0, 1)$, meaning that the generator does not always generate correct / incorrect action proposals; otherwise, the performance will be deterministic and a verifier would not be needed. 

\paragraph{Theorem A.} Given $p_G \in (0, 1)$, $N > 1$, then $\mathbb{E}\left[R_{\text{gt}}(a_{\text{w/ver}})\right] > \mathbb{E}\left[R_{\text{gt}}(a_{\text{naive}})\right] \iff p_V > 0.5$

\textit{Proof. } 

The expected reward of naive sampling from generator: $\mathbb{E}\left[R_{\text{gt}}(a_{\text{naive}})\right] = p_G \times 1 + (1-p_G) \times 0 = p_G$.

The expected reward of the selected action using the verifier is: 
\begin{alignat*}{2}
\mathbb{E}[R_{\text{gt}}(a_{\text{w/ver}})] 
&= P(\exists i,\ V(a_i) = 1) \cdot \mathbb{E}[R_{\text{gt}}(a) \mid V(a) = 1] \\
&\quad + P(\forall i,\ V(a_i) = 0) \cdot \mathbb{E}[R_{\text{gt}}(a) \mid V(a) = 0] \\
&= \left(1 - (1 - Q)^N\right) \cdot \frac{P(R = 1, V = 1)}{P(V = 1)} \\
&\quad + (1 - Q)^N \cdot \frac{P(R = 1, V = 0)}{P(V = 0)}
\end{alignat*}

$\text{where } Q = P\left(V(a)=1\right) = (1-p_G)(1-p_V) + p_Gp_V$. 
and
\[
P(R = 1, V = 1) = p_G \cdot p_{V}, \quad P(R = 1, V = 0) = p_G \cdot (1 - p_{V}).
\]

Substituting into the expression, we get:
\[
\mathbb{E}[R_{\text{gt}}(a_{\text{w/ver}})] 
= \left(1 - (1 - Q)^N\right) \cdot \frac{p_G p_{V}}{Q} 
+ (1 - Q)^N \cdot \frac{p_G (1 - p_{V})}{1 - Q}.
\]

We will first prove $\mathbb{E}[R_{\text{gt}}(a_{\text{w/ver}})] > \mathbb{E}[R_{\text{gt}}(a_{\text{naive}})] \Rightarrow p_{V} > 0.5$, and show that each step is reversible to prove the other direction.

\begin{align*}
&\mathbb{E}[R_{\text{gt}}(a_{\text{w/ver}})] > \mathbb{E}[R_{\text{gt}}(a_{\text{naive}})] = p_G. \\
\overset{divide \ by}{\underset{p_G}{\iff}}
&\left(1 - (1 - Q)^N\right) \cdot \frac{p_{V}}{Q} + (1 - Q)^N \cdot \frac{1 - p_{V}}{1 - Q} > 1.
\end{align*}

Rewriting and simplifying:
\begin{align*}
&\left(1 - (1 - Q)^N\right) \cdot \frac{p_{V}}{Q} + (1 - Q)^N \cdot \frac{1 - p_{V}}{1 - Q} > 1 \\
\iff
&\left(1 - (1 - Q)^N\right) \cdot \frac{p_{V}}{Q} + (1 - p_{V})(1 - Q)^{N-1} > 1 \\
\iff
&\left(\frac{p_{V}}{Q} - \frac{p_{V}}{Q}(1 - Q)^N\right) + (1 - p_{V})(1 - Q)^{N-1} > 1 \\
\iff
&\frac{p_{V}}{Q} - \frac{p_{V}}{Q}(1 - Q)^N + (1 - p_{V})(1 - Q)^{N-1} > 1 \\
\iff
&\left(\frac{p_{V}}{Q} - 1\right) + \left[- \frac{p_{V}}{Q}(1 - Q)^N + (1 - p_{V})(1 - Q)^{N-1}\right] > 0 \\
\iff
&\left(\frac{p_{V}}{Q} - 1\right) + (1 - Q)^{N-1} \left[- \frac{p_{V}}{Q}(1 - Q) + (1 - p_{V})\right] > 0 \\
\iff
&\left(\frac{p_{V}}{Q} - 1\right) + (1 - Q)^{N-1} \left[1 - p_{V} - \frac{p_{V}}{Q}(1 - Q)\right] > 0 \\
\iff
&\left(\frac{p_{V}}{Q} - 1\right) + (1 - Q)^{N-1} \left(1 - \frac{p_{V}}{Q}\right) > 0 \\
\iff
&\left(\frac{p_{V}}{Q} - 1\right)\left[1 - (1 - Q)^{N-1}\right] > 0.
\end{align*}

Since $p_G\in(0, 1)$, we have \( Q \in (0, 1) \), then \( 1 - (1 - Q)^{N-1} > 0 \), so the inequality holds if and only if:
\[
\frac{p_{V}}{Q} > 1 \quad \iff \quad p_{V} > Q.
\]

Substituting the expression for \( Q \):
\[
p_{V} > p_G p_{V} + (1 - p_G) (1-p_{V}),
\]

Rearranging:
\begin{align*}
p_{V} - p_G p_{V} &> (1 - p_G) (1-p_{V}), \\
\iff
p_{V}(1 - p_G) &> (1 - p_G) (1-p_{V}).
\end{align*}

Since \( p_G \in (0,1) \), we can divide both sides by \( 1 - p_G \), yielding:
\[
p_{V} > 1 - p_{V} \quad \iff \quad p_V > 0.5.
\]

Since all of the above steps are reversible, we prove that the verifier improves the expected reward over naive sampling if and only if \( p_{V} > 0.5 \) .

Therefore, when the verifier has better than random accuracy $p_V > 0.5$, the overall expected reward is improved. This theory demonstrates the effect of our generator-verifier system.

\begin{figure}[h]
    \centering
    \includegraphics[width=1.0\textwidth]{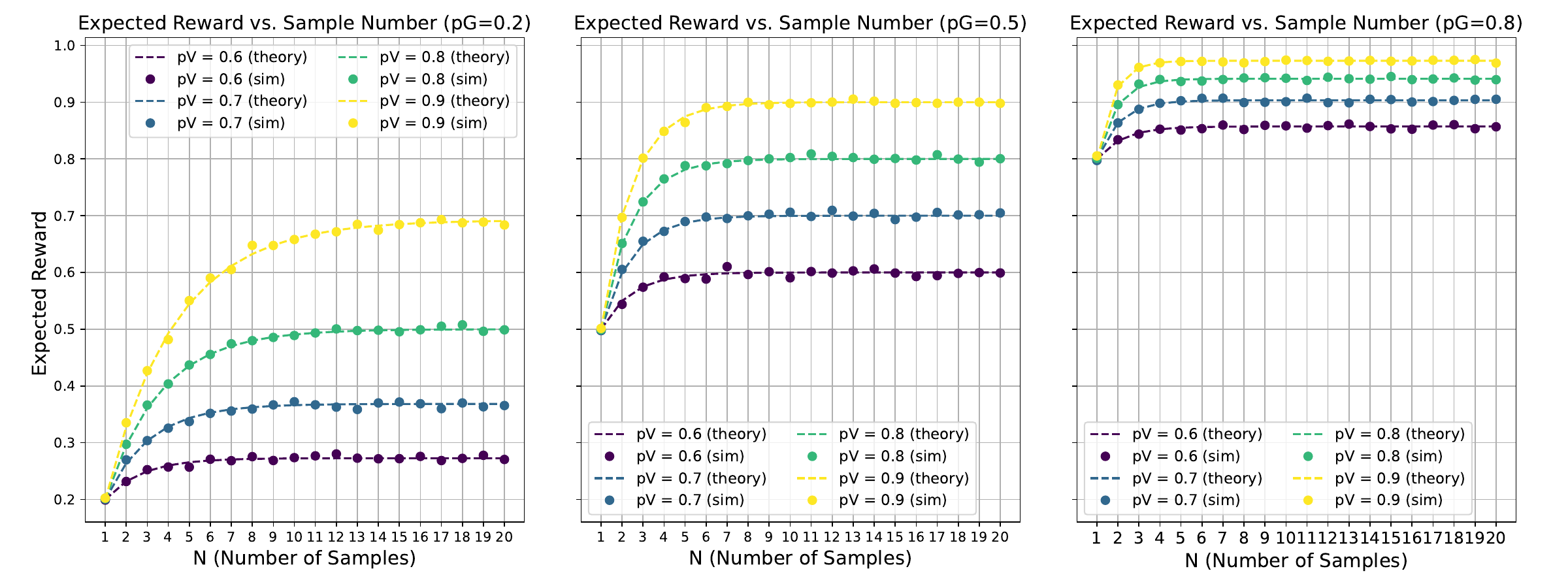} 
    \caption{\textbf{Expected and Simulated Reward w.r.t. Sample Number}: We can see the expected and simulated improvements the verifier introduces with different generator and verifier accuracy.}
    \label{fig:expected_reward}
\end{figure}

\paragraph{Numerical Analysis:} We also demonstrate the expected reward as a function of the number of samples $N$ for different values of $p_G$ and $p_V$ in Fig.~\ref{fig:expected_reward}. We plot the simulated results where we simulate 10k trials, within each trial we sample $N$ samples and simulate the ground truth and verifier reward with $p_G$ and $p_V$. We can see even a very biased generator ($p_G=0.2$) can be significantly improved by a good verifier ($p_V=0.9$) to almost 3x accuracy with only 10 samples.

\subsection{Discrete Reward with Dependent Generator Verifier Accuracies}
\label{sec:theory_discrete_dependent}

We can also extend the theory to cases where the generator and the verifier is not independent. In the binary reward setting, dependence implies that the accuracy of the verifier differs depending on whether the generated action is good or bad, that is, \( p_{V1} = P(V(a) = R_{gt}(a) \mid R_{\text{gt}}(a) = 1) \) and \( p_{V0} = P(V(a) = R_{gt}(a) \mid R_{\text{gt}}(a) = 0) \). Below we show our theory can be extended to this case.

Let \( a \sim \pi_G \) be an action sampled from the generator, $p_G$ be the probability of the generated action having reward 1.  
Let \( R_{\text{gt}}(a) \in \{0, 1\} \) be the ground-truth reward (i.e., whether the action is good or bad), and \( V(a) \in \{0, 1\} \) be the verifier output.

We define:
\begin{alignat*}{2}
&p_G = P\bigl(R_{\text{gt}}(a)=1 \,\bigm|\, a \sim \pi_G\bigr)\\
&p_{V1} = P\left(V(a) = R_{\text{gt}}(a) =1 \mid R_{\text{gt}}(a) = 1\right)\\
&p_{V0} = P\left(V(a) = R_{\text{gt}}(a) = 0 \mid R_{\text{gt}}(a) = 0\right)
\end{alignat*}
that is, \( p_{G} \) is the probability the generator generates a good action, and \textbf{the verifier's accuracy is dependent on the generation quality} (ground truth reward of the action), \( p_{V1} \) is the probability the verifier predicts 1 given the action is good, and \( p_{V0} \) is the probability the verifier predicts 0 given the action is bad.

\paragraph{Thm. A$^\prime$.} Given $p_G \in (0, 1), N > 1$, then $\mathbb{E}[R_{\text{gt}}(a_{\text{w/ver}})] > \mathbb{E}[R_{\text{gt}}(a_{\text{naive}})] \iff p_{V1} + p_{V0} > 1$.

\textit{Proof.} 

The expected reward of naive sampling from generator: $\mathbb{E}\left[R_{\text{gt}}(a_{\text{naive}})\right] = p_G \times 1 + (1-p_G) \times 0 = p_G$

The expected reward of the selected action using the verifier is:
\begin{alignat*}{2}
\mathbb{E}[R_{\text{gt}}(a_{\text{w/ver}})] 
&= P(\exists i,\ V(a_i) = 1) \cdot \mathbb{E}[R_{\text{gt}}(a) \mid V(a) = 1] \\
&\quad + P(\forall i,\ V(a_i) = 0) \cdot \mathbb{E}[R_{\text{gt}}(a) \mid V(a) = 0] \\
&= \left(1 - (1 - Q)^N\right) \cdot \frac{P(R = 1, V = 1)}{P(V = 1)} \\
&\quad + (1 - Q)^N \cdot \frac{P(R = 1, V = 0)}{P(V = 0)}
\end{alignat*}

where
\[
Q = P(V(a) = 1) = p_G \cdot p_{V1} + (1 - p_G) \cdot (1 - p_{V0}),
\]
and
\[
P(R = 1, V = 1) = p_G \cdot p_{V1}, \quad P(R = 1, V = 0) = p_G \cdot (1 - p_{V1}).
\]

Substituting into the expression, we get:

\[
\mathbb{E}[R_{\text{gt}}(a_{\text{w/ver}})] 
= \left(1 - (1 - Q)^N\right) \cdot \frac{p_G p_{V1}}{Q} 
+ (1 - Q)^N \cdot \frac{p_G (1 - p_{V1})}{1 - Q}.
\]

We will first prove $\mathbb{E}[R_{\text{gt}}(a_{\text{w/ver}})] > \mathbb{E}[R_{\text{gt}}(a_{\text{naive}})] \Rightarrow p_{V1} + p_{V0} > 1$, and show that each step is reversible to prove the other direction.

\begin{align*}
&\mathbb{E}[R_{\text{gt}}(a_{\text{w/ver}})] > \mathbb{E}[R_{\text{gt}}(a_{\text{naive}})] = p_G. \\
\overset{divide \ by}{\underset{p_G}{\iff}}
&\left(1 - (1 - Q)^N\right) \cdot \frac{p_{V1}}{Q} + (1 - Q)^N \cdot \frac{1 - p_{V1}}{1 - Q} > 1.
\end{align*}

Rewriting and simplifying (omitting steps similar with proof of Theory A):
\begin{align*}
&\left(1 - (1 - Q)^N\right) \cdot \frac{p_{V1}}{Q} + (1 - Q)^N \cdot \frac{1 - p_{V1}}{1 - Q} > 1 \\
\iff
&\left(1 - (1 - Q)^N\right) \cdot \frac{p_{V1}}{Q} + (1 - p_{V1})(1 - Q)^{N-1} > 1 \\
\iff
&\left(\frac{p_{V1}}{Q} - \frac{p_{V1}}{Q}(1 - Q)^N\right) + (1 - p_{V1})(1 - Q)^{N-1} > 1 \\
\iff
&\frac{p_{V1}}{Q} - \frac{p_{V1}}{Q}(1 - Q)^N + (1 - p_{V1})(1 - Q)^{N-1} > 1 \\
\iff
&\left(\frac{p_{V1}}{Q} - 1\right)\left[1 - (1 - Q)^{N-1}\right] > 0.
\end{align*}

Since $p_G\in(0, 1)$, we have \( Q \in (0, 1) \), then \( 1 - (1 - Q)^{N-1} > 0 \), so the inequality holds if and only if:
\[
\frac{p_{V1}}{Q} > 1 \quad \iff \quad p_{V1} > Q.
\]

Substituting the expression for \( Q \):
\[
p_{V1} > p_G p_{V1} + (1 - p_G) (1-p_{V0}),
\]

Rearranging:
\begin{align*}
p_{V1} - p_G p_{V1} &> (1 - p_G) (1-p_{V0}), \\
\iff
p_{V1}(1 - p_G) &> (1 - p_G) (1-p_{V0}).
\end{align*}

Since \( p_G \in (0,1) \), we can divide both sides by \( 1 - p_G \), yielding:
\[
p_{V1} > 1 - p_{V0} \quad \iff \quad p_{V1} + p_{V0} > 1.
\]

Since all of the above steps are reversible, we prove that the verifier improves the expected reward over naive sampling if and only if \( p_{V1} + p_{V0} > 1 \) .

\subsection{Continuous Reward}
\label{sec:theory_continuous}

We now extend the theoretical analysis to continuous rewards. Consider a task where the ground truth reward function $R_{gt}:A\to \mathbb{R}$ that assigns real-valued rewards. The generator $G$ produces actions $a\in A$ such that $R_{gt}(a)$ follows a distribution with mean $\mu_G$ and variance $\sigma^2_{G}$. The verifier $V:A \to \mathbb{R}$ provides scores for a given action. We make the following assumptions:

\begin{itemize}
\item Assumption 1: The generator's reward follows a normal distribution $X=R_{gt}(a) \sim \mathcal{N}(\mu_G, \sigma_G^2)$.
\item Assumption 2: The verifier's score satisfies: $Y = V(a) = X(a) + \epsilon$, where $\epsilon \sim \mathcal{N}(0, \sigma_V^2)$ and is independent of $X$, therefore $Y\sim \mathcal{N}(\mu_G, \sigma_G^2+\sigma_V^2)$.
\end{itemize}

We compare the expected reward of 2 methods:
\begin{itemize}
    \item \textbf{Method 1 (Naive Sampling):} Sample 1 action $a_{\text{naive}} \sim G$, the expected reward is $\mathbb{E}\left[R_{\text{gt}}(a_{\text{naive}})\right] = \mu_G$.
    
     \item \textbf{Method 2 (Verifier Selection):}  
    Sampling \( N \) actions and selecting \( a_{\text{w/ver}} = \arg\max_i V(a_i) \). We denote the expected reward as $\mathbb{E}[R_{gt}(a_{\text{w/ver}})] = \mu_G + \Delta_{ver}$. 
\end{itemize}

We denote $X_{(N)}$ as the maximum ground truth reward of the $N$ samples: $X_{(N)} = \max\{X_1, X_2, ..., X_N\}$, where $X_i = R_{gt}(a_i)$.  We also define $Y_{(N)}$ as the maximum verifier's reward of the $N$ samples: $Y_{(N)} = \max\{Y_1, Y_2, ..., Y_N\}$, where $Y_i = V(a_i)$. 


Following results derived from Extreme Value Theory on normal variables (Theorem 1.5.3 in \citet{leadbetter2012extremes} and Theorem 5.1 from \citet{dasgupta2014sharp}), we first approximate $\mathbb{E}[Z_{(N)}]$ where $Z_{(N)}$ is the maximum reward of $N$ samples from a standard normal distribution $\mathcal{N}(0, 1)$:

\vspace{-6pt}
$$\mathbb{E}[Z_{(N)}] \approx \sqrt{2\ln{N}}-\frac{\ln\ln{N}+\ln(4\pi)}{2\sqrt{2\ln{N}}} + \frac{\gamma}{\sqrt{2\ln{N}}}$$
$$\mathbb{E}[Y_{(N)}] = \mu_G+\sqrt{\sigma_G^2+\sigma_V^2}\cdot \mathbb{E}[Z_{(N)}]$$

Given Assumptions 1 and 2, $X$ and $Y$ obey a bivariate normal distribution with correlation $\rho$:
\[
\begin{aligned}
\mathrm{Cov}(X,Y)
&= \mathbb{E}\bigl[X\,(X+\epsilon)\bigr]\;-\;\mathbb{E}[X]\,\mathbb{E}[X+\epsilon] \\[6pt]
&= \bigl(\mathbb{E}[X^2]+\underbrace{\mathbb{E}[X\,\epsilon]}_{=\,\mathbb{E}[X]\mathbb{E}[\epsilon]=0}\bigr)
   \;-\;\mathbb{E}[X]\bigl(\mathbb{E}[X]+\underbrace{\mathbb{E}[\epsilon]}_{=0}\bigr) \\[6pt]
&= \mathbb{E}[X^2]-\mathbb{E}[X]^2 \;=\;\sigma_G^2.
\end{aligned}
\]
$$\rho=\frac{Cov(X,Y)}{\sigma_X, \sigma_Y} = \frac{\sigma_G^2}{\sigma_G \cdot \sqrt{\sigma_G^2 + \sigma_V^2}}
=\frac{\sigma_G}{\sqrt{\sigma_G^2+\sigma_V^2}}
$$

Therefore the conditional expectation for X given Y is $\mathbb{E}[X|Y=y] = \mu_X+\sigma_X\,\rho\,\frac{y-\mu_Y}{\sigma_Y} = \mu_G + \sigma_G^2/(\sigma_G^2 + \sigma_V^2)*(y-\mu_G)$, Therefore we have:


\[
\begin{aligned}
\Delta_{\rm ver}
&=\;\mathbb{E}_{Y_{(N)}}\bigl[\;\mathbb{E}[\,X\mid Y=Y_{(N)}]\bigr]\;-\,\mu_G
&&\text{(law of total expectation over }Y_{(N)})\\
&=\;\frac{\sigma_G^2}{\sigma_G^2+\sigma_V^2}\,\Bigl(\mathbb{E}[\,Y_{(N)}\,]-\mu_G\Bigr)
&&\text{(substitute }E[X\mid Y=y]\text{)}\\
&=\;\frac{\sigma_G^2}{\sigma_G^2+\sigma_V^2}\;\mathbb{E}[\,Z_{(N)}\,]\sqrt{\sigma_G^2+\sigma_V^2}\\
&=\;\mathbb{E}[\,Z_{(N)}\,]\;\frac{\sigma_G^2}{\sqrt{\sigma_G^2+\sigma_V^2}}\,.
\end{aligned}
\]

\begin{figure}[h]
    \centering
    \includegraphics[width=1.0\textwidth]{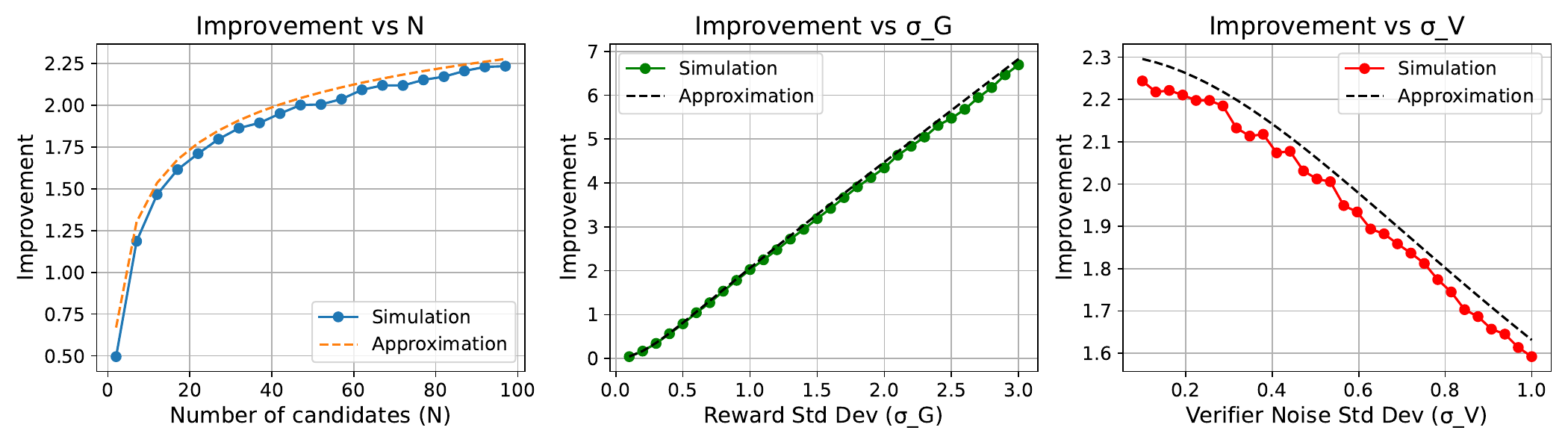} 
    \caption{\textbf{Simulation Results (Normal Reward)}: We plot the simulated improvements and the theoretical improvements against different $N$, $\sigma_G$, and $\sigma_V$. The default values for parameters that are fixed are $N = 50$, $\sigma_G = 1.0$, and $\sigma_V=0.5$.}
    \label{fig:theory_sim_normal}
\end{figure}

We provide numerical simulation results to visualize and validate the approximation: at each trial, we sample N samples from $\mathcal{N}(\mu_G, \sigma^2_G)$, and add a verifier noise to each sample $\epsilon \sim \mathcal{N}(0, \sigma^2_V)$, and record the difference (improvement) of the reward of the sample with the largest verifier score and the reward of the first sample. Then we plot the simulated average improvement over 10k trials with the theoretical approximation against different $N$, $\sigma_G$, and $\sigma_V$ in Fig.~\ref{fig:theory_sim_normal}. We can see the theoretical approximation stays close to the simulation data and is intuitively interpretable. As \(N\) grows, \(E[Z_{(N)}]\sim\sqrt{2\ln N}\), so the gain increases with the number of samples, in proportion to \(\sqrt{2\ln N}\).  For fixed verifier noise \(\sigma_V\), the prefactor \(\sigma_G^2/\sqrt{\sigma_G^2+\sigma_V^2}\) grows nearly linearly in \(\sigma_G\) once \(\sigma_G\gg\sigma_V\), meaning a wider ground‐truth reward spread yields almost proportional extra potential benefit from the verifier. Conversely, as the verifier uncertainty \(\sigma_V\) increases, the same prefactor decays like \(1/\sigma_V\) meaning that the improvement brought by verifier decays inversely proportional to \(\sigma_V\).  

\begin{figure}[h]
    \centering
    \includegraphics[width=1.0\textwidth]{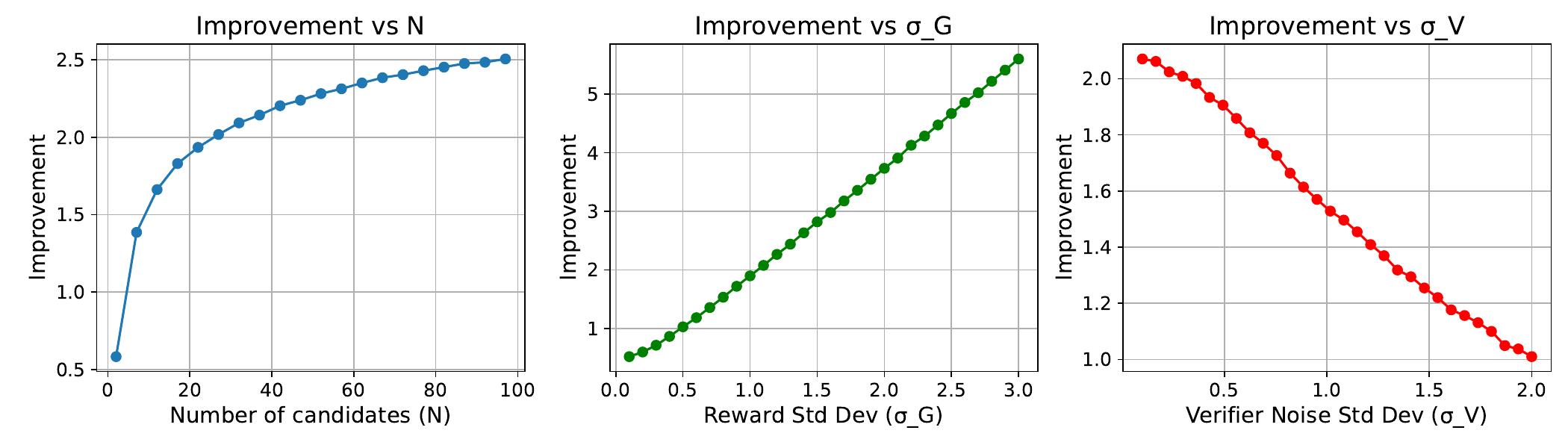} 
    \caption{\textbf{Simulation Results (Gaussian Mixture Model Reward)}: We plot the simulated improvements and the theoretical improvements against different $N$, $\sigma_G$ and $\sigma_V$. The default values for parameters that are fixed are $N = 20$, $\sigma_G=1.0$ and $\sigma_V=0.5$.}
    \label{fig:theory_sim_gmm}
\end{figure}

We also numerically simulate improvements with different continuous reward distribution and verifier model apart from the bivariate normal definition above: 

\paragraph{Gaussian Mixture Reward} We simulate the situation where the reward of the generator's proposals follows a gaussian mixture model distribution $p(x) = 0.5 * \mathcal{N}(-0.5, \sigma^2_G) + 0.5 * \mathcal{N}(0.5, \sigma_G^2)$ (e.g. when the generator can generate multiple action modes with equal possibilities, like push and pull a door). We plot the improvements against $N$, $\sigma_G$ and $\sigma_V$ in Fig.~\ref{fig:theory_sim_gmm}.

\begin{figure}[h]
    \centering
    \includegraphics[width=0.8\textwidth]{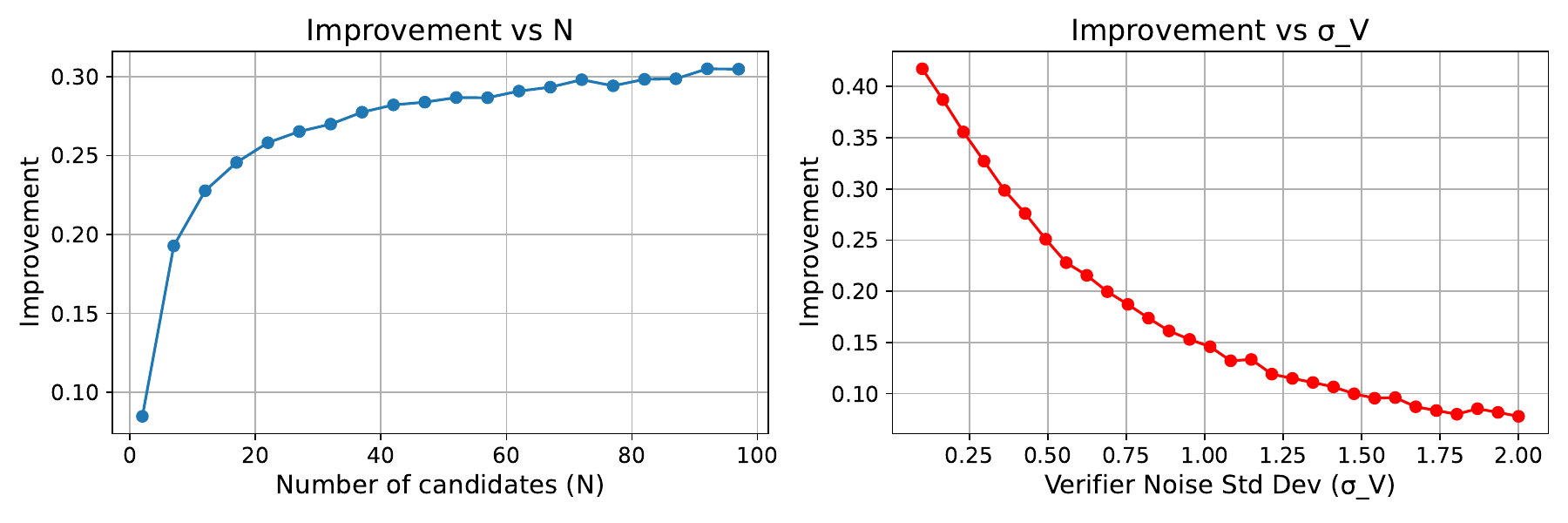} 
    \caption{\textbf{Simulation Results (Uniform Reward)}: We plot the simulated improvements and the theoretical improvements against different $N$, and $\sigma_V$. The default values for parameters that are fixed are $N = 20$ and $\sigma_V=0.5$.}
    \label{fig:theory_sim_uniform}
\end{figure}

\paragraph{Uniform Reward} We simulate the situation where the reward of the generator's proposals follows a uniform distribution $U(0, 1)$ (e.g. when the generator can generate various actions with equal possibilities), and plot the improvements against $N$ and $\sigma_V$ in Fig.~\ref{fig:theory_sim_uniform}. We can see that the with about 10 samples the improvement is beginning to become significant.

\begin{figure}[h]
    \centering
    \includegraphics[width=0.8\textwidth]{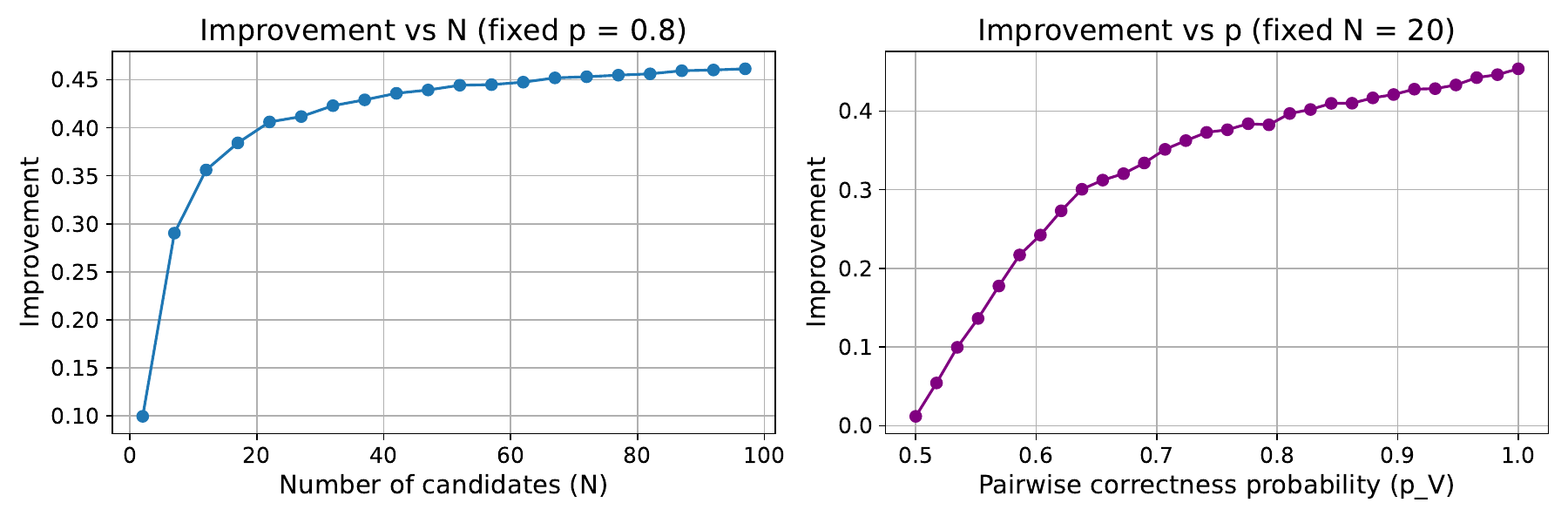} 
    \caption{\textbf{Simulation Results (w.r.t. Verifier Pairwise Accuracy)}: We plot the simulated improvements and the theoretical improvements against different $N$, and $p_V$. The default values for parameters that are fixed are $N = 20$ and $p_V=0.8$.}
    \label{fig:theory_sim_pairwise}
\end{figure}

\paragraph{Verifier Pairwise Accuracy} We also demonstrate the effect of verifier against verifier pairwise accuracy: $P\left(V(a) > V(b) \ | \ R_{gt}(a) > R_{gt}(b)\right) = p_V$. This measure removes the assumption that the verifier is an unbiased estimation of the ground truth reward. We simulate with reward distribution of $U(0, 1)$ and plot the improvements against $N$ and $p_V$ in Fig.~\ref{fig:theory_sim_pairwise}. We can see that with $p_V > 0.6$ there is a significant improvement.

Using the above theoretical and numerical analysis, we see the improvements of the method that uses the verifier together with the generator (compared to only using the generator) across various reward distributions, both discrete and continuous, validating the soundness and efficacy of our generator-verifier approach.

\section{Method Details}
\label{sec:model-details}


\subsection{Action Generator}

\begin{wrapfigure}{r}{0.5\textwidth}   
  \vspace{-10pt}                         
  \centering
  \includegraphics[width=\linewidth]{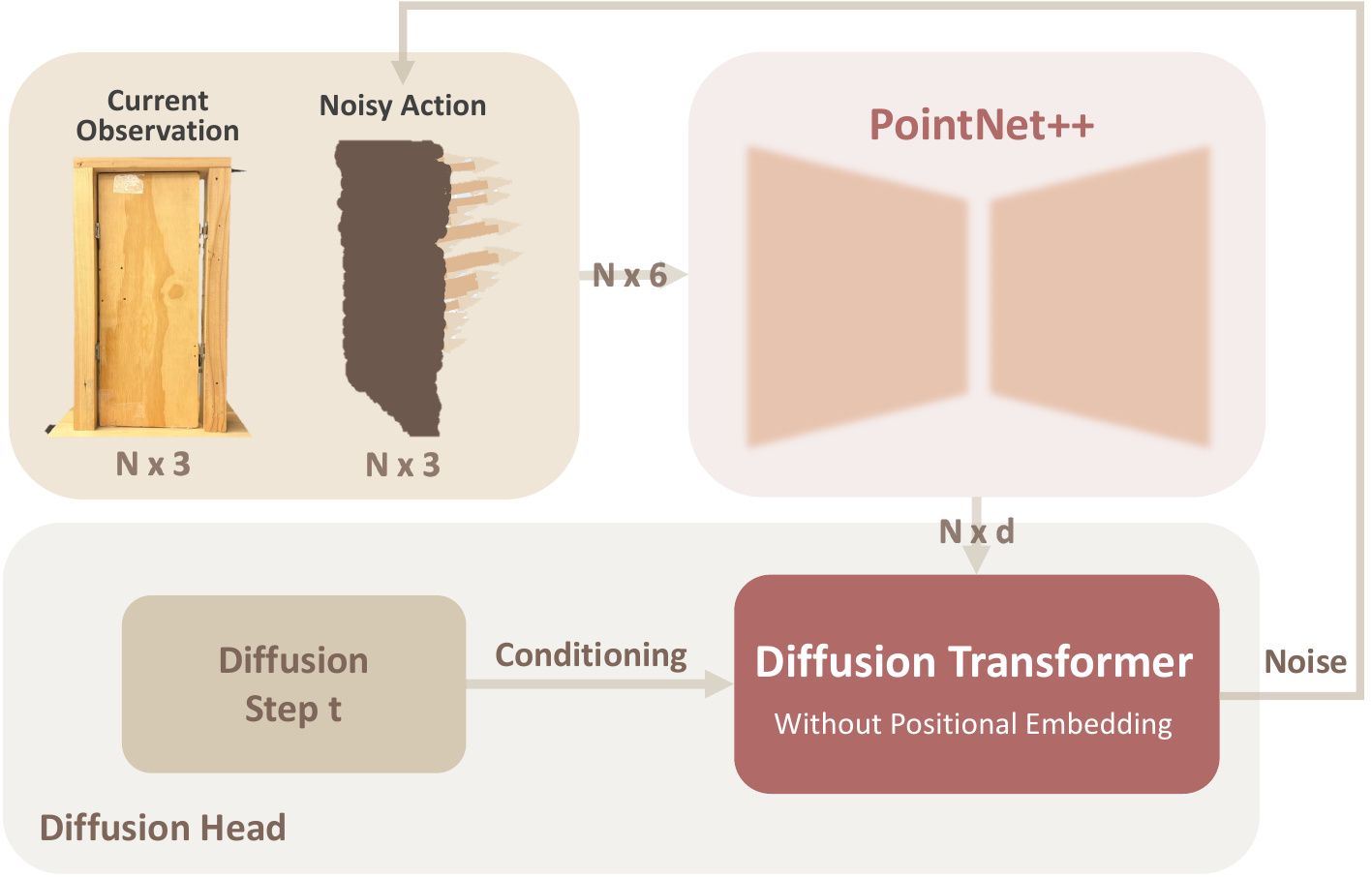}
  \caption{\textbf{Action Generator Architecture}}
  \label{fig:action_generator}
\end{wrapfigure}

Following \citet{li2024flowbothd}, we construct our action generator with a PointNet++~\cite{qi2017pn2} encoder $E_{\Theta}$ and a Diffusion Transformer (DiT)~\cite{peebles2023scalable} denoiser $D_{\Phi}$ as shown in Fig.~\ref{fig:action_generator}. We remove the positional encoding in original DiT because point clouds are unordered. Given the current observation point cloud $o \in \mathbb{R}^{N \times 3}$, and a noisy dense 3D articulation flow (defined in ~\citet{eisner2022flowbot3d}) $a_{noisy} \in \mathbb{R}^{N \times 3}$, we first use the PointNet++ encoder to encode the noisy action into a 
point-wise embedding ($\mathbb{R}^{N\times d}$, where b is the embedding dimension). Then we input the per-point features into a Diffusion Transformer (DiT) to predict the noise $\epsilon$ to denoise from the noisy action field. 

In our setting, we use $N = 1200$, $d = 128$. And the size of DiT is depth = 5, hidden size = 128, and 4 heads. We train with learning rate = 1e-4, AdamW optimizer with weight decay = 1e-5, batch size = 32 and train for 400 epochs. 

\paragraph{From 3DAF to Dense Action Field} The generator generates 3DAF to provide flexibility of selecting grasp points, and we by default choose the point with the largest magnitude as the action point (which means the largest articulation leverage as illustrated in ~\citet{eisner2022flowbot3d}), and the corresponding flow as the action direction. The action is then transformed into a dense action field for verifier evaluation. The verifier takes in dense action field because of (1) generating random actions for training is more intuitive compared with generating random flows that needs specification of an articulation, (2) action is a more generalizable concept and representation.

For the uneven object pick-up task, we cannot directly apply articulation flow representation due to the lack of hinge mechanism for articulated objects. Instead, our generator represents each surface point by a “pick‑up flow” vector: its direction is normal to the object’s surface (vertical), and its magnitude equals the point’s distance from the object’s center of mass. We then define leverage analogously, selecting the action point as the location with the smallest pick-up flow magnitude.

\subsection{Dataset Details}
\label{sec:dataset_details}

We sample 200 history trajectories for each object in the dataset with various length (range of 1 to 30 for articulated objects and 1 to 10 for uneven object pick-up), and for each trajectory, we include 10 random actions along with ground truth actions and transformed history actions (actions are transformed to the current state, e.g. if the history action is based on closed doors while the current state is an open door, the history action point and direction will be transformed to the corresponding open angle). We simulate each action to collect corresponding ground truth reward labels to supervise training. 

\paragraph{Ground truth rewards}
The ground truth reward are defined according to task goal: for opening articulated objects, the reward is defined as the relative amount the action opens the door against the ground-truth action at the same state; for uneven object lifting, the reward is the signed distance between the lift point and the mass center.

\subsection{Model Details}

For action and action result sequential reasoning module, we each use 4 layers of self-attention with 4 heads, and the action and action result embedding dimensions are both 128. We sample 1200 points for each point cloud during training and inference. We train the scoring module for 50 epochs (or early stop at convergence) using AdamW optimizer with lr = 1e-4.

\section{Baseline and Ablation Details}
\label{sec:baselines-details}

In this section, we describe the architecture of our baselines and ablations in details.

\subsection{Oracle Experiments}
\label{sec:oracle_details}

We include two types of oracle experiments for better understanding of the model's performance: (1) Oracle Sampler: We always append the ground truth action into the action proposals; this experiment aims to test the verifier's ability with optimal generator; (2) Oracle Verifier: We always choose the closest-to-ground-truth action from the batch when selecting from the action proposals; this experiment aims to test the upper limit of using a verifier on the current generator. 

\subsection{Sparse Action}

In the \textit{Sparse Action} ablation, instead of representing an action as a dense action field (Sec. 5.3), we represent the action as a sparse action field where only the contact point has nonzero values: given a point cloud $\{p_i\}$, and an action point $p$ and direction $d$, the sparse action field is the point cloud with 3 extra channels to represent the action, calculated as $\{d_i: d *\mathds{1}(p_i=p)\}$. We keep the exact same architecture, but force the first downsampling layer to sample the contact point to prevent complete information loss due to downsampling.

\subsection{Point Cloud as Result}
\label{sec:pcd_as_result}

\begin{figure}[h]
    \centering
    \includegraphics[width=1.0\textwidth]{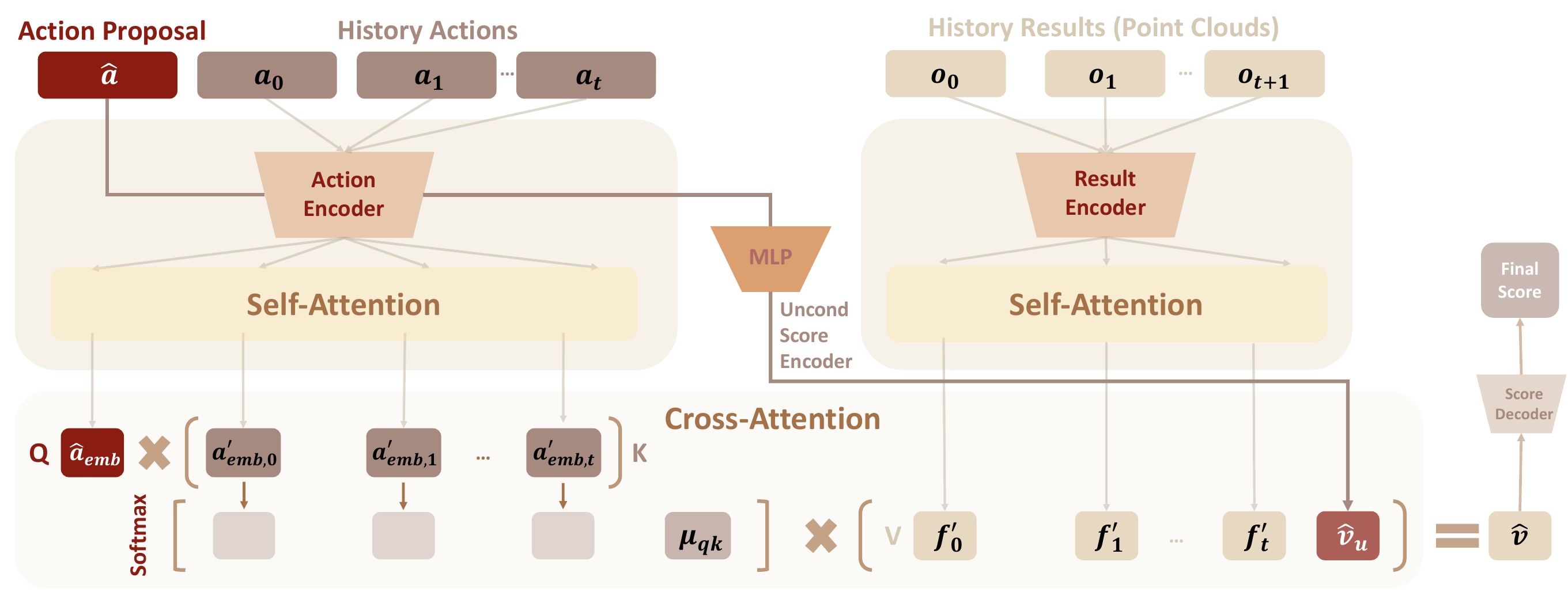} 
    \caption{\textbf{Point Cloud as Result (Ablation) Architecture}: We pass the observations sequence directly through a self-attention layer without extracting observation flow from observation pairs.}
    \label{fig:pcd_as_result}
\end{figure}

In the \textit{Point Cloud as Result} ablation, instead of explicitly extracting observation flow from the observation pairs before and after an action, we directly encode and pass the observations sequence to a self-attention layer to extract corresponding features as shown in Fig.~\ref{fig:pcd_as_result}. This gives the model all the information needed to complete the task. The model would need to understand the movements between frames implicitly, unlike our architecture that explicitly computes the flow between observations. 
The other parts of the architecture are kept the same as HAVE.

\subsection{Vanilla Transformer}
\label{sec:vanilla_transformer_arch}

\begin{figure}[h]
    \centering
    \includegraphics[width=0.8\textwidth]{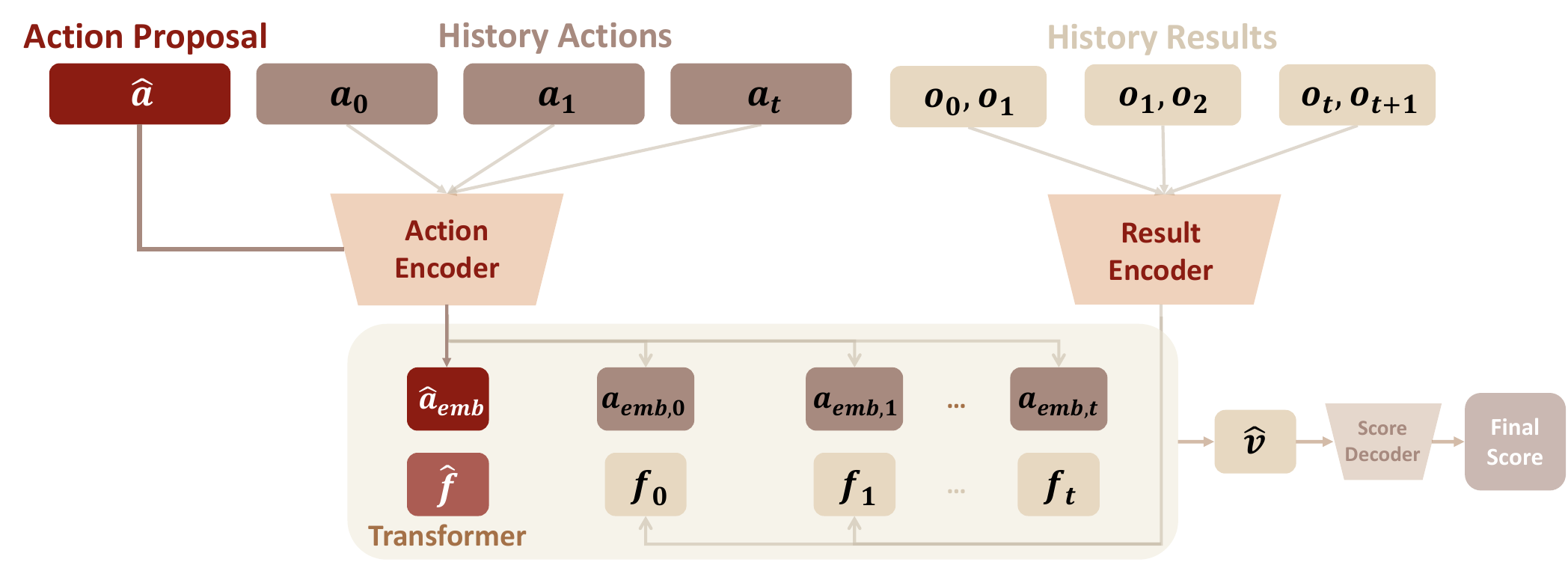} 
    \caption{\textbf{Vanilla Transformer (Ablation) Architecture}: We concatenate the PointNet++ encoded action features and result features and pass through an end-to-end transformer.}
    \label{fig:vanilla_transformer}
\end{figure}

In the \textit{Vanilla Transformer} ablation in Sec. 6.3, instead of encoding actions and results independently and using an explicit dot-product attention layer to reason about the action proposal, in this ablation we concatenate the encoded action features with corresponding result features, and use an end-to-end transformer to predict the final score as shown in Fig.~\ref{fig:vanilla_transformer}. We first encode action proposal $\hat{a}$, history actions and results $\{(o_{i-1}, a_i, o_i), i \in [t-1]\}$ in the same way as HAVE: $\hat{a}_{\textrm{emb}} = E_{\textrm{action}}(\hat{a})$, $\{a_{\textrm{emb},i} = E_{\textrm{action}}(a_i), i\in[t-1]\}$, $\{f_i = E_{\textrm{obs}}(o_{i-1}, o_i), i\in[t-1]\}$. Then we concatenate the corresponding actions and results $\{h_i = concat(a_{\textrm{obs}}, f_i), i\in[t-1]\}$. We concatenate the action proposal embedding with an learnable token $\hat{f}$ to keep the same dimension with $\{h_i\}$: $\hat{h} = concat(\hat{a}_{\textrm{emb}}, \hat{f})$. The sequence $\{\hat{h}, h_0, h_1, ...,h_t\}$ is then passed through a transformer that produce embedding tokens for each timestep: $\{\hat{h}', h_0', h_1' ..., h_t'\}$, we take $\hat{v} = \hat{h}'$ as the final score embedding for the action proposal, and pass through an MLP to obtain the final score prediction. 

We supervise the model with 2 losses, final score loss that compares the predicted score with the ground truth score for the action proposal, and history score loss in an encoder-decoder manner where we induce another MLP to decode the result features $\{f_i\}$ to scores and supervise with the ground truth scores for history actions to encourage learning better representation of action results.

\subsection{Conditional Diffusion}
\label{sec:cond_diff}

\begin{figure}[h]
    \centering
    \includegraphics[width=1.0\textwidth]{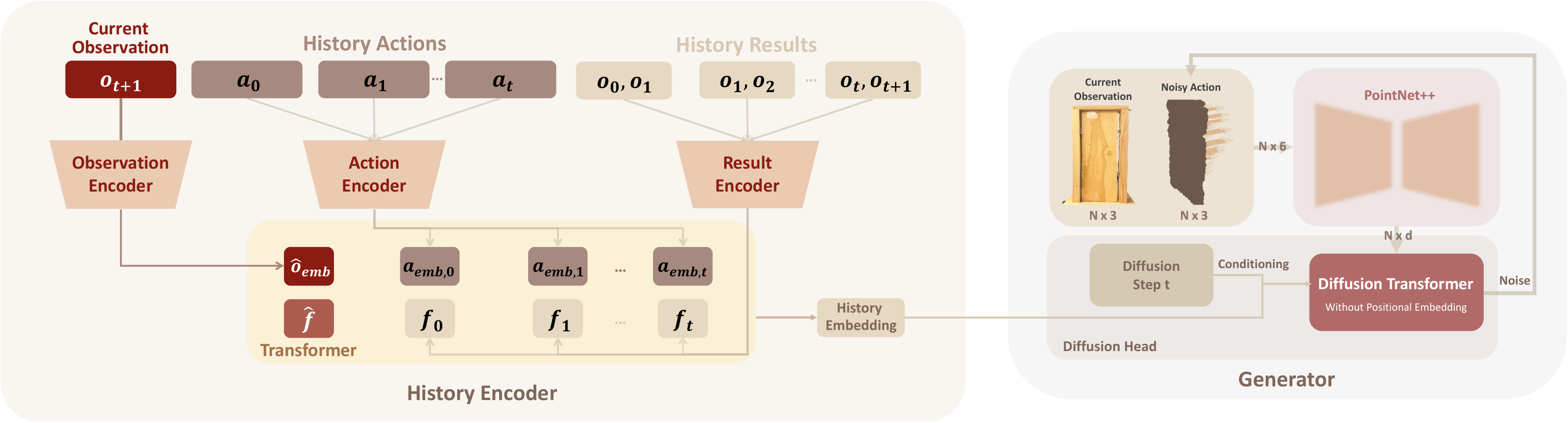} 
    \caption{\textbf{Conditional Diffusion Architecture}: We first embed the histories and inject the embedding as a condition into DiT in the denoiser.}
    \label{fig:cond_diffusion}
\end{figure}

To compare HAVE with end-to-end history-aware conditional diffusion without verifier, we construct a conditional diffusion structure merging the unconditional generator and the history-aware verifier structure as shown Fig.~\ref{fig:cond_diffusion}. We separate the history-aware generator into a history encoder, and a generator head. The history encoder takes in current observation / state, and the
history actions and results, and encodes everything into a history embedding representing the guidance information obtained from history interactions. We use the vanilla transformer architecture in Appendix D.3 because instead of outputting a score for a specific action proposal, the history embedding should include information about the possible action itself where the HAVE architecture focused on calculating a score is not directly applicable. 

The history embedding will be added with diffusion timestep as a condition into the Diffusion Transformer denoiser through adaLN-Zero conditioning. We train the model with classifier-free guidance style where we randomly drop out histories during training. When there is no history, the history embedding will be replaced by a learnable token. 

\section{Experiment Results and Details}

\subsection{Held-out Articulated Categories}
\label{sec: heldout_results}

\begin{figure}[h]
    \centering
    \includegraphics[width=0.8\textwidth]{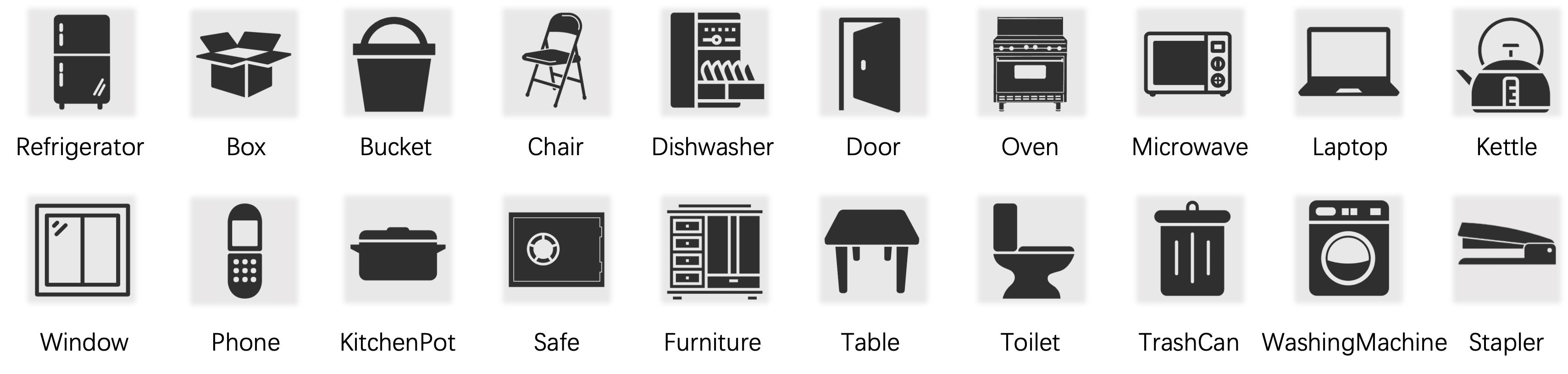} 
    \caption{\textbf{Articulated Objects Categories Legend}: We show the icon-to-text correspondence for better interpretation of the table.}
    \label{fig:articulated_table_icons}
\end{figure}

Following the protocol of \citet{eisner2022flowbot3d}, we split PartNet-Mobility dataset~\cite{xiang2020sapien} into 11 training categories, and leave 10 categories entirely unseen.  In Table~\ref{tab:tvu-failure-test}, we report the failure rate on unseen categories for each of the baselines and for our method. We provide the icon-to-text correspondence in Fig.~\ref{fig:articulated_table_icons} for easier understanding of the corresponding categories. We observe significant performance improvements with our method compared with baselines without a verifier. We underperform FlowBotHD (w/CC) where it includes a heuristic that filters out actions that doesn't obey with successful history action movements. This consistency check in FlowBotHD is like a task-specific ground truth verifier, and due to its heuristic design, it is not directly generalizable to other tasks outside of articulated objects (e.g. in uneven objects pick-up, the important phase is to reason about failures where there is no successful history available). Instead, our verifier architecture provides a generalizable insight and achieves comparable results. For ablations, we observe removing unconditional score loss slightly improves the performance. This indicates that whether or not supervise the unconditional score is a minor design choice whose performance may vary across tasks.

\input{tables/fr_train-val-unseen_unseen}

\subsection{Uneven Object Pick-up}
\label{sec:uneven_object_details}

\begin{figure}[h]
    \centering
    \includegraphics[width=0.8\textwidth]{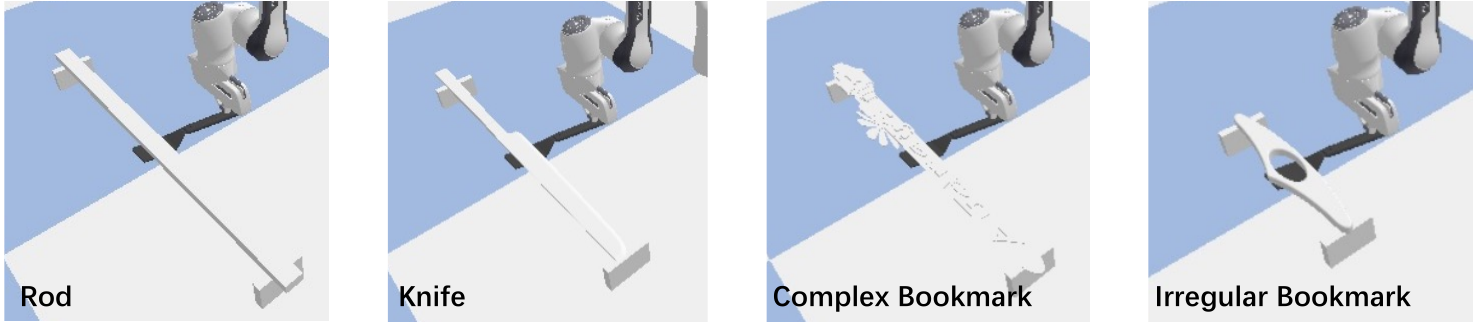} 
    \caption{\textbf{Uneven Object Dataset Examples}: From left to right, we show examples of the visual appearance of rod, knife, complex bookmark and irregular bookmark in our uneven object dataset.}
    \label{fig:uneven_object_example}
\end{figure}

We constructed the training ambiguous rod dataset with 17 rod replicates of the same size and assigned the mass center randomly across the rod length. For the unseen test dataset, we included 40 rods (with sizes and mass centers different from training dataset), 20 complex bookmarks, 20 irregular bookmarks, and 20 knives of various lengths, widths, and heights. Examples of the visual appearances of each categories are shown in Fig.~\ref{fig:uneven_object_example}.

\subsection{Real-World Ambiguous Door}
\label{sec:real_world_details}

\begin{figure}[h]
    \centering
    \includegraphics[width=1.0\textwidth]{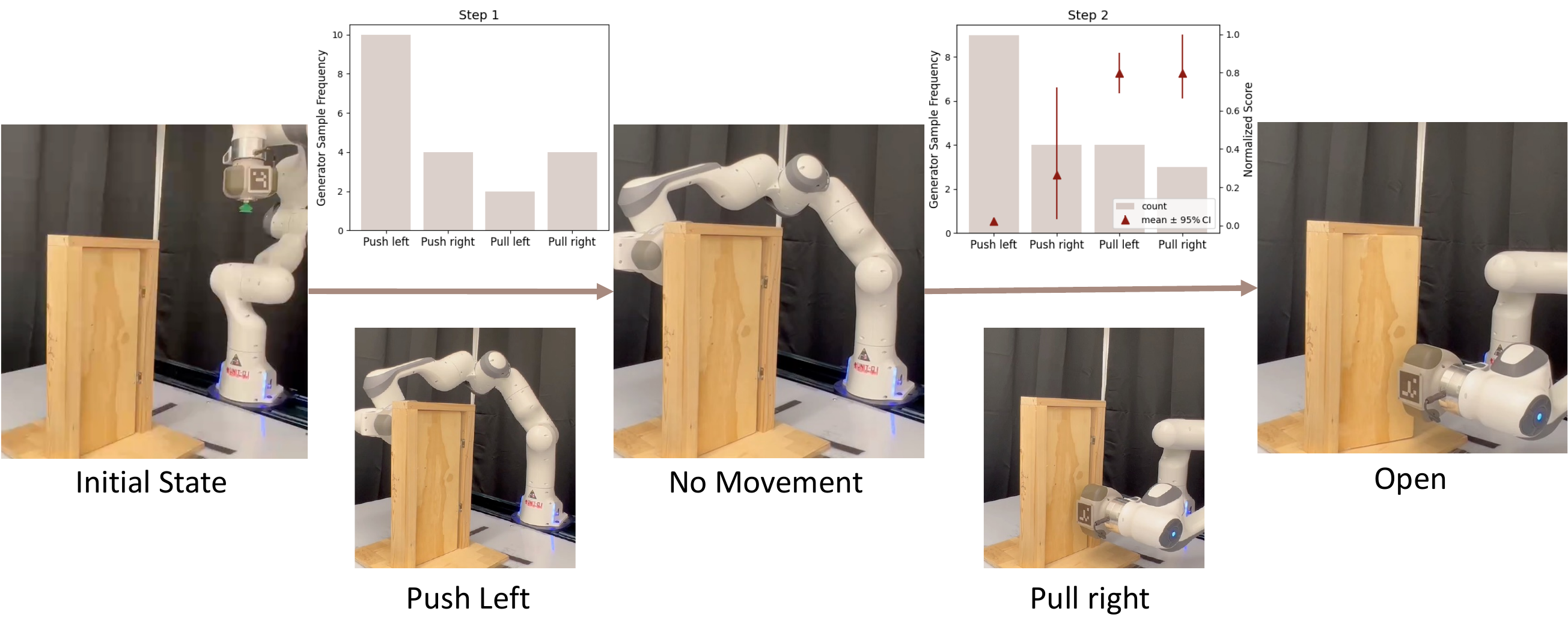} 
    \caption{\textbf{Real World Analysis (Example 1)}: HAVE suppresses the failure mode “push left” despite it being frequently generated from the generator and opens the door.}
    \label{fig:real_world_analysis_1}
\end{figure}

To show HAVE's ability to transfer to real world scenarios, we visualize the process of two real world trials. We plot the sampling distribution from the generator, the average verifier score for each mode, and the 95\% confidence interval to demonstrate how the generator and the verifier work together. In Fig.~\ref{fig:real_world_analysis_1}, the door is configured to be opened by pull right (opens from the front right), and the model first tries push left (push is also executed by pull from the back) which does not move the door. For the next step, the sampled batch from the generator has a similar distribution because the observation is similar. Despite being frequently generated by the generator, the failure mode (push left) receives consistently low scores (low mean and very small confidence interval). The verifier ends up choosing the correct action (pull right) among the three remaining actions and opens the door. 

\begin{figure}[h]
    \centering
    \includegraphics[width=1.0\textwidth]{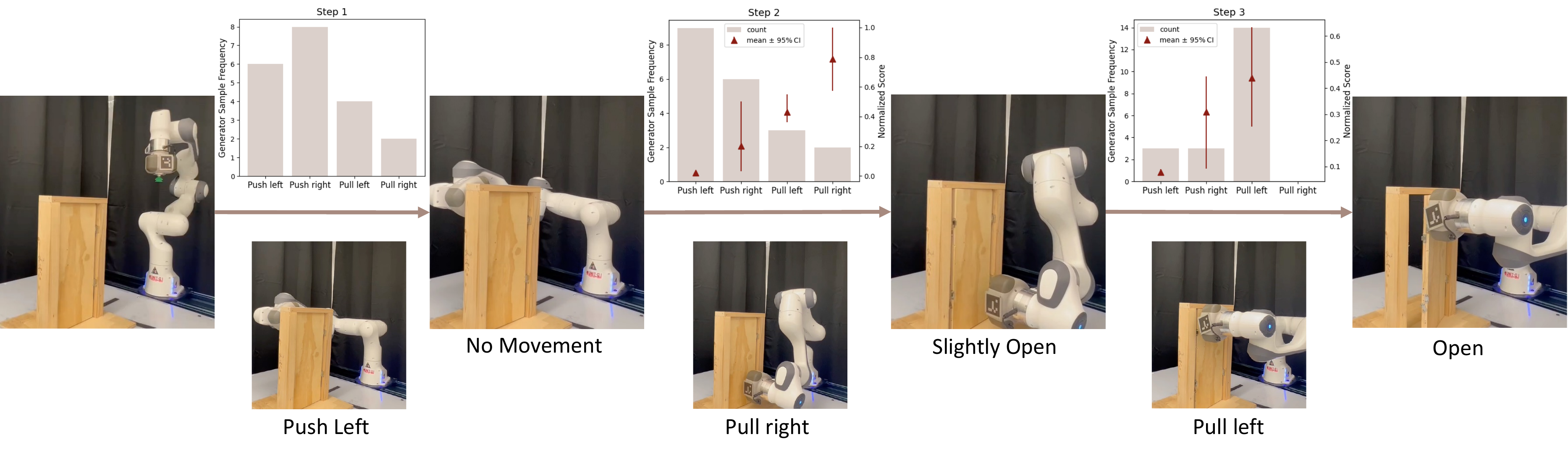} 
    \caption{\textbf{Real World Analysis (Example 2)}: HAVE takes 3 steps to explore towards the correct open action without repeating failure modes.}
    \label{fig:real_world_analysis_2}
\end{figure}

In Fig.~\ref{fig:real_world_analysis_2}, the door is configured to open by pull left. The first two steps are similar, except that the pull right action opens the door for a small amount. Then given the slightly opened door, the generator is also biased towards the correct mode based on the geometry, and the verifier assigns higher scores for actions that are not failures before, together results in the correct open mode. This demonstrates that with verifier and generator working together, we are effectively utilizing both geometric and history cues towards a successful decision.

\subsection{Generation-Verification for Hierarchical Policy Learning}
\begin{figure}[h]
    \centering

    \begin{subfigure}[b]{0.3\textwidth}
        \centering
        \begin{minipage}{0.48\textwidth}
        \includegraphics[width=\linewidth]{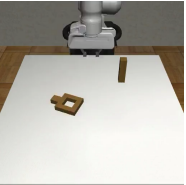}
        \end{minipage}
        \hfill
        \begin{minipage}{0.48\textwidth}
        \includegraphics[width=\linewidth]{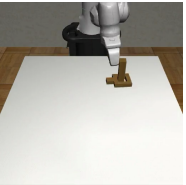}
        \end{minipage}
        \caption{Square D2}
    \end{subfigure}
    \hfill
    \begin{subfigure}[b]{0.3\textwidth}
        \centering
        \begin{minipage}{0.48\textwidth}
        \includegraphics[width=\linewidth]{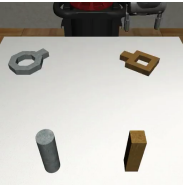}
        \end{minipage}
        \hfill
        \begin{minipage}{0.48\textwidth}
        \includegraphics[width=\linewidth]{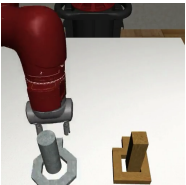}
        \end{minipage}
        \caption{Nut Assembly D0}
    \end{subfigure}
    \hfill
    \begin{subfigure}[b]{0.3\textwidth}
        \centering
        \begin{minipage}{0.48\textwidth}
        \includegraphics[width=\linewidth]{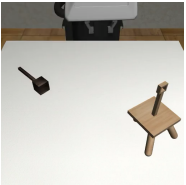}
        \end{minipage}
        \hfill
        \begin{minipage}{0.48\textwidth}
        \includegraphics[width=\linewidth]{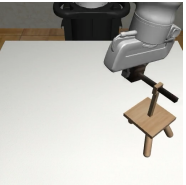}
        \end{minipage}
        \caption{Threading D2}
    \end{subfigure}
    \caption{\textbf{Experimental environments from MimicGen}~\cite{mandlekar2023mimicgen}: The left image in each sub figure shows an initial state of the environment, and the right image shows the goal state.}
    \label{fig:mimicgen}
\end{figure}

Our idea of using a verifier in ambiguous settings can also be extended to other settings like hierarchical policy learning pipeline. In this section, we utilize high-level sub-goal priors to guide low-level policies to tackle benchmark robotic manipulation tasks in MimicGen~\cite{mandlekar2023mimicgen} (Figure~\ref{fig:mimicgen}). 

\textbf{Sub-goal Generation for Multi-stage Tasks.} We consider manipulation tasks in MimicGen~\cite{mandlekar2023mimicgen} using a robot manipulator with 2-finger grippers. Given an expert demo trajectory $\tau=(s_0, a_0, s_1, a_1, \cdots, s_T)$, we define critical sub-goal frames $t_1, t_2,\cdots, t_k\in \{0, 1, \cdots, T-1\}$ as the time steps when the robot gripper switches from open state to closed state or from closed to open. For each time step between two critical sub-goal frames in the expert trajectory, i.e., $t\in [t_i, t_{i+1})$, we augment the data point by incorporating the robot gripper point cloud at the sub-goal frame $t_{i+1}$ as the ground-truth sub-goal information for our hierarchical pipeline, i.e.,
\begin{equation}
    \mathcal{D}=\{(s_t, a_t, P_{t_{i+1}})|t\in [t_i, t_{i+1})\}_{t=0}^{T-1},
\end{equation}
where $P_{t_{i+1}}$ is the robot gripper point cloud at time step $t_{i+1}$.

\textbf{Hierarchical System Pipeline.} The pipeline mainly consists of two critical parts: (a) \textit{High-level sub-goal predictor} takes visual observation as input and predicts the next sub-goal in the form of gripper point clouds. (b) \textit{Low-level robot policy} takes visual observation and sub-goal conditioning as input and generates low-level robot action through diffusion process. 

For high-level sub-goal generator, we follow the model backbone in TAX3D~\cite{cai2024tax3d}. In training time, the input of the model is current gripper point cloud and the full-scene point cloud. The ground-truth target flow is the per-point displacement between the current gripper point cloud and the sub-goal gripper point cloud. For each expert demo trajectory $\tau=(s_0, a_0, s_1, a_1,\cdots, s_T)$, we create $T$ input-label paired data points for training the TAX3D model. During inference time, randomly sampled displacements $\Delta X\sim N(0, I)$ are de-noised conditioned on current gripper features and full-scene features. The final $\Delta X_0$ is predicted to displace the gripper into a sub-goal configuration. 

For the low-level policy, we use a modified version of 3D Diffusion Policy (DP3)~\cite{Ze2024DP3}, which consists of a point cloud encoder that encodes the point cloud observation into a latent embedding, and a conditional denoising diffusion model that generates noise-free actions for execution. For the point cloud encoder, we perform cross attention among full-scene point cloud, current gripper points and sub-goal gripper points with Rotary Position Embedding~\cite{su2024roformer}.

\textbf{Verifier as Sub-goal Selector.} In this setting, high-level sub-goal prediction features implicit multi-modality and ambiguity, due to the fact that the TAX3D model inputs are very similar near critical sub-goal frames while target predictions switch from the previous sub-goal to the next stage. Take Square D2 (Figure~\ref{fig:mimicgen}(a)) as an example, when the gripper is about to grasp the object, the high-level prediction of sub-goal gripper points should switch from the object handle to the target peg. To this end, we use a separately-trained verifier to select the best sub-goal prediction from multiple TAX3D-generated high-level predictions.

More specifically, we use PointNet++~\cite{qi2017pn2} as the model backbone and fit the model for binary classification. In training time, we feed the network with positive demos that consist of full-scene points and ground-truth sub-goal gripper points, and negative demos that consist of full-scene points and randomly perturbed ground-truth sub-goal gripper points. In inference time, we use the model to score TAX3D samples by passing in current full-scene points and predicted gripper points, and then choose the one with the highest score as the high-level sub-goal prediction.

\textbf{Results.} We report the performance of the whole hierarchical pipeline with different high-level conditions in Table~\ref{Table:mimicgen}. \textit{TAX3D} refers to conditioning on TAX3D predictions with 1 sample, and \textit{TAX3D+Verifier} refers to sampling 5 TAX3D predictions and using the score predictor to select the best sample as high-level guidance in inference. Generally, the verifier helps tackle ambiguity at critical goal-switching frames and guides the low-level policy to better complete benchmark tasks. This experiment supports our main idea of using a verifier for better action selection and demontrates the insight's generalization across tasks.
\input{tables/mimicgen}

\section{Analysis}

\subsection{Verifier Performance w.r.t History Type}

\begin{figure*}[ht]
    \centering
    \begin{minipage}[c]{0.3\textwidth}
        \centering
        \vspace{10pt}
        \resizebox{\textwidth}{!}{
        \begin{tabular}{|r|c|}
            \hline
            \textbf{History Length} & \textbf{Valid Rate} \\
            \hline
            One step & 99.79\%\\
            Two steps & 97.47\% \\
            Three steps & 99.01\% \\
            \hline
        \end{tabular}
        }
        \captionof{table}{\textbf{Failure History Analysis (Valid Rate)}: the top-selected actions almost always avoid repeating failures, indicating that failed actions are suppressed in later selections.}
        \label{tab:failure_analysis_table}
    \end{minipage}
    \hfill
    \begin{minipage}[c]{0.68\textwidth}
        \centering
        \includegraphics[width=1.0\textwidth]{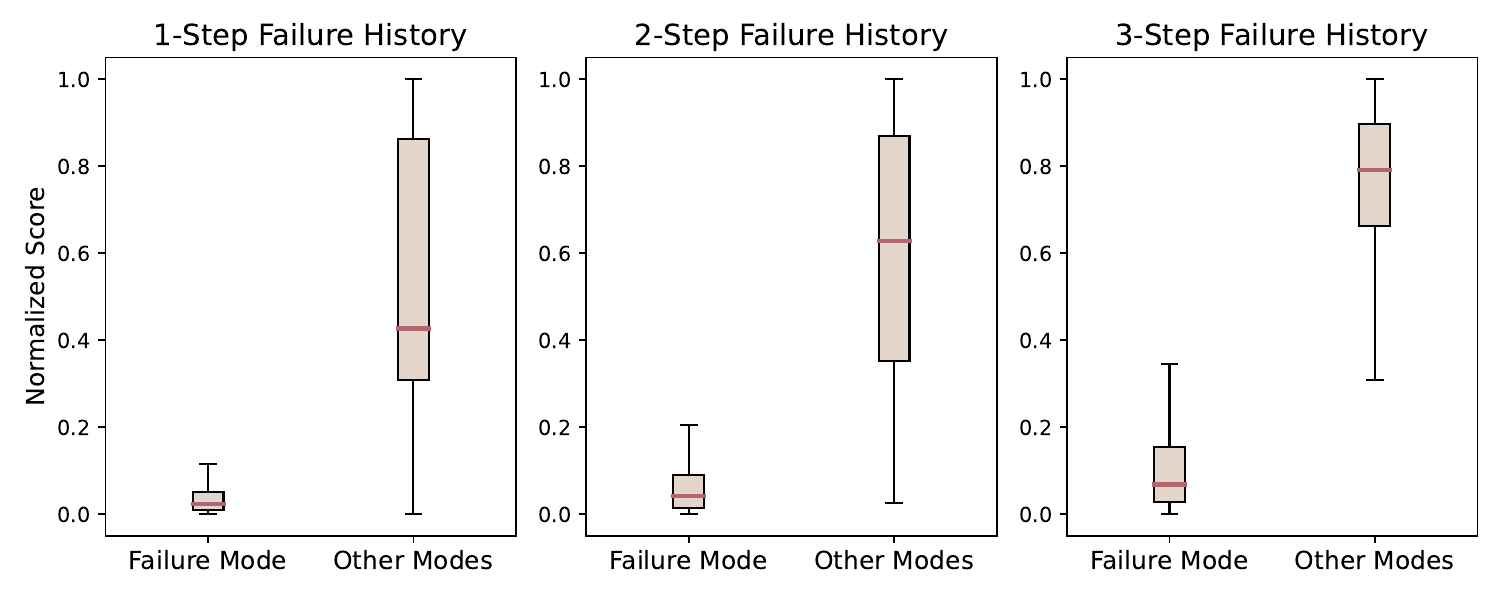} 
        \caption{\textbf{Failure History Analysis (Average Scores)}: Verifier score analysis under different failure histories. Actions matching failure modes receive lower scores, showing the verifier’s ability to learn from failure histories.}
        \label{fig:failure_history_analysis}
    \end{minipage}
\end{figure*}

\paragraph{Failure History (Articulated objects)}: On the multi-modal door dataset, we create failure interaction histories (observation flow is set to zeros to represent failure results) with different lengths and failure modes and visualize the normalized scores predicted by the verifier in Fig.~\ref{fig:failure_history_analysis}. We sample 20 actions (different contact point) from each mode (push left, push right, pull left, pull right). Given a specific failure history sequence, we evaluate the sampled actions with the verifier and calculate the normalized score, if the corresponding mode of the action appears in the history sequence, it will be counted as the ``failure modes", and other actions are counted into ``other modes". 

We plot the distribution of normalized score (normalizedS to [0,1] range using min-max normalization for better visualization) predicted for failure modes and other modes across different history actions, and different doors in the multi-modal door dataset. We can see that on average the predicted scores for failure modes are lower than other modes. We also demonstrate in Table~\ref{tab:failure_analysis_table} the percentage of the trials where the selected action (action with the highest score) among the 20 * 4 proposed actions lies in ``other modes". We observe almost 100\% of the chosen actions avoid failure actions, demonstrating HAVE's ability to understand failures and avoid repeating failures.

\begin{figure}[h]
    \centering
    \includegraphics[width=1.0\textwidth]{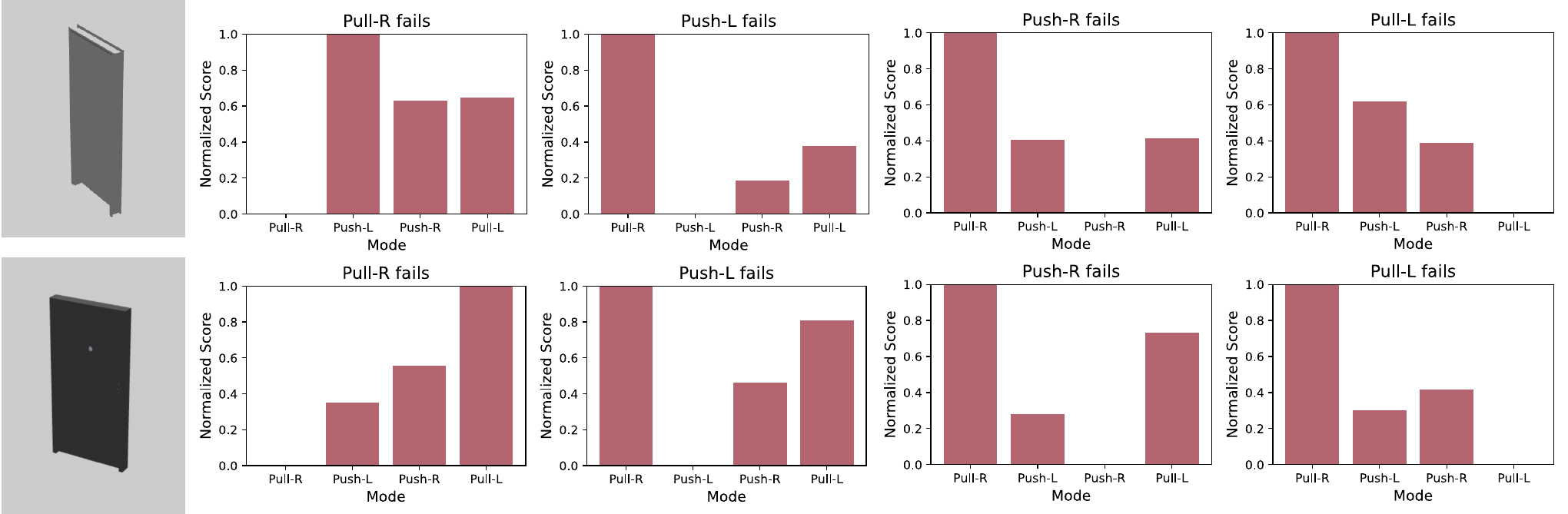} 
    \caption{\textbf{Failure History Analysis (Examples)}: Verifier predictions for each mode under one-step failure history on two example doors, showing score suppression of previously failed modes.}
    \label{fig:failure_history_analysis_exp}
\end{figure}

We also demonstrate detailed score predictions for each mode given one step failure history for 2 randomly selected doors in Fig.~\ref{fig:failure_history_analysis_exp}. We can see that the mode that has led to failure in the history is suppressed by the verifier. 

\begin{figure}[h]
    \centering
    \includegraphics[width=1.0\textwidth]{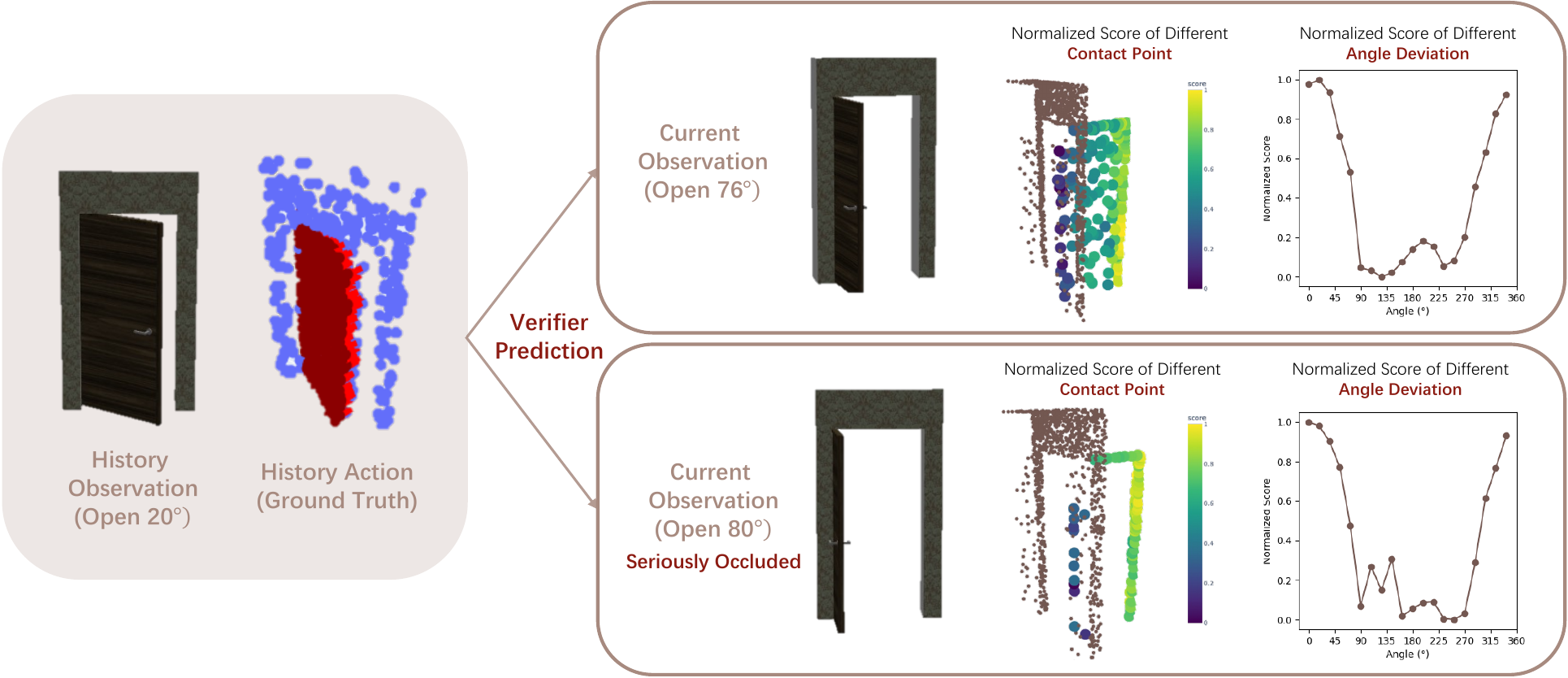} 
    \caption{\textbf{Successful History Analysis}: Given a successful history, we use the verifier to evaluate action proposals on a new state (geometrically very different from the history observation) of the same object. (1) With a fixed ground-truth direction, the score heatmap for varying contact points shows the verifier favors points with better leverage; (2) With a fixed optimal contact point, the verifier scores different action directions, correctly assigning higher scores near the ground truth. 
    }
    \label{fig:success_history_exp}
\end{figure}

\paragraph{Success History (Articulated Objects)}: To test the verifier's ability to utilize success histories, we demonstrate the verifier's evaluation for actions under seriously occluded point clouds given a successful history action. In Fig.~\ref{fig:success_history_exp}, we give the verifier a successful history at 20 degrees, and make it reason about occluded point clouds at around $80^\circ$. We first construct action batch with the correct action direction but different action points, and plot the normalized score heatmap. We can see that the verifier basically understands what is a better grasp point (larger leverage) despite the point clouds are geometrically ambiguous. We then keep the action point the same (as the top leverage point in ground truth), and rotate the ground truth action direction around the z-axis (vertical to the ground), and plot their predicted normalized scores. We can see the scores are higher when it is closer to the ground truth direction (closer to $0^{\circ}$ and $360^{\circ}$). This experiment demonstrates the verifier's ability to refer to previous successful histories (even with large open angle difference and thus very different point cloud) and reason about seriously occluded states with fairly precise understanding of good action points and directions.

\begin{figure}[h]
    \centering
    \includegraphics[width=1.0\textwidth]{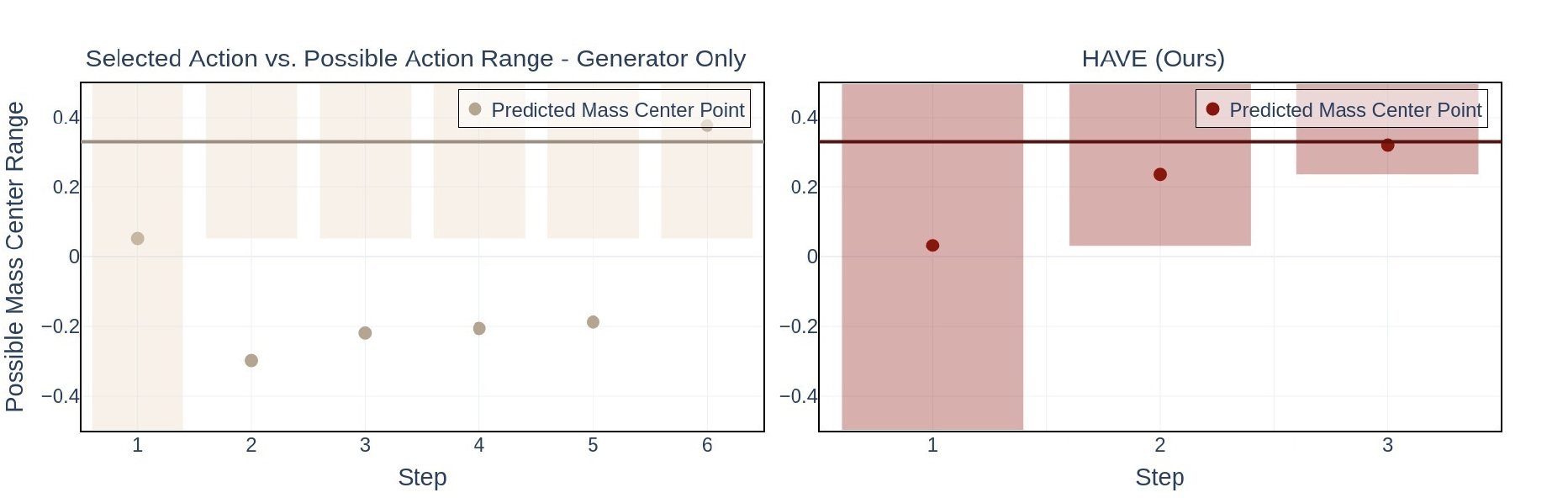} 
    \caption{\textbf{Uneven Object Verifier Analysis (Example 1)}: HAVE narrows the action space each step by leveraging information from failures, leading to a faster success. }
    \label{fig:failure_analysis_uo_1}
\end{figure}

\paragraph{Failure History (Uneven Objects)} We also demonstrate a similar learning from failure mechanism for uneven object pick-up. During each pick-up attempt, the tilting direction of the object will leak information about which side the center of mass lies on. Therefore, we can calculate a theoretical center of mass range at each step based on the previous histories. Like in Fig.~\ref{fig:uneven-object-fig}, we plot the ground truth center of mass as a horizontal line, the theoretical center of mass range as bars, and the action point chosen at each step as scattered dots. The ideal behavior is that with history information, the model can choose within the theoretical boundaries at each step and converge to the ground truth position with fewer steps. In Fig.~\ref{fig:failure_analysis_uo_1}, we can see that HAVE's choices lie within the theoretical range at each step, and uses only half of the steps used by the generator. In comparison, the generator-only method repeatedly fails regardless of valuable information from the history failures.  

\begin{figure}[h]
    \centering
    \includegraphics[width=1.0\textwidth]{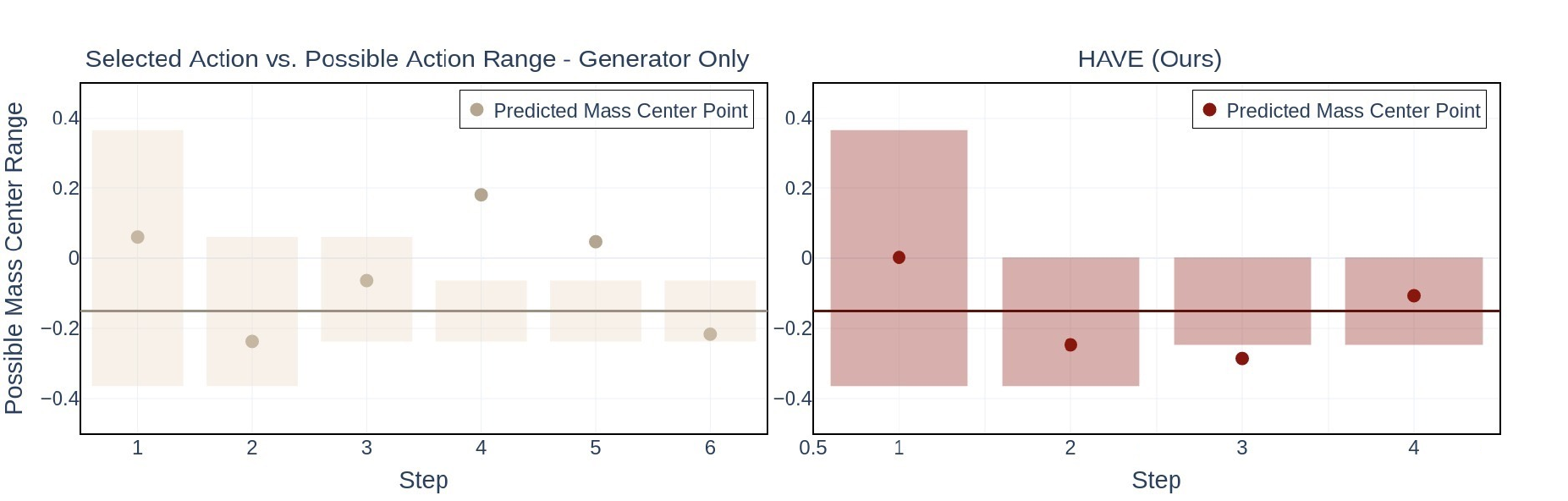} 
    \caption{\textbf{Uneven Object Verifier Analysis (Example 2)}: HAVE learns from history, its selected actions stay mostly within the theoretical range at each step.}
    \label{fig:failure_analysis_uo_2}
\end{figure}

Similarly in Fig.~\ref{fig:failure_analysis_uo_2}, the generator-only method breaks out of the theoretical boundary more often, while ours mostly lies within the boundary with the third step being a slight violation. The visualizations demonstrate HAVE's ability to actually reason about history actions and extract useful information from previous failures.

\subsection{Verifier Performance w.r.t History Length}

\begin{figure}[h]
    \centering
    \includegraphics[width=1.0\textwidth]{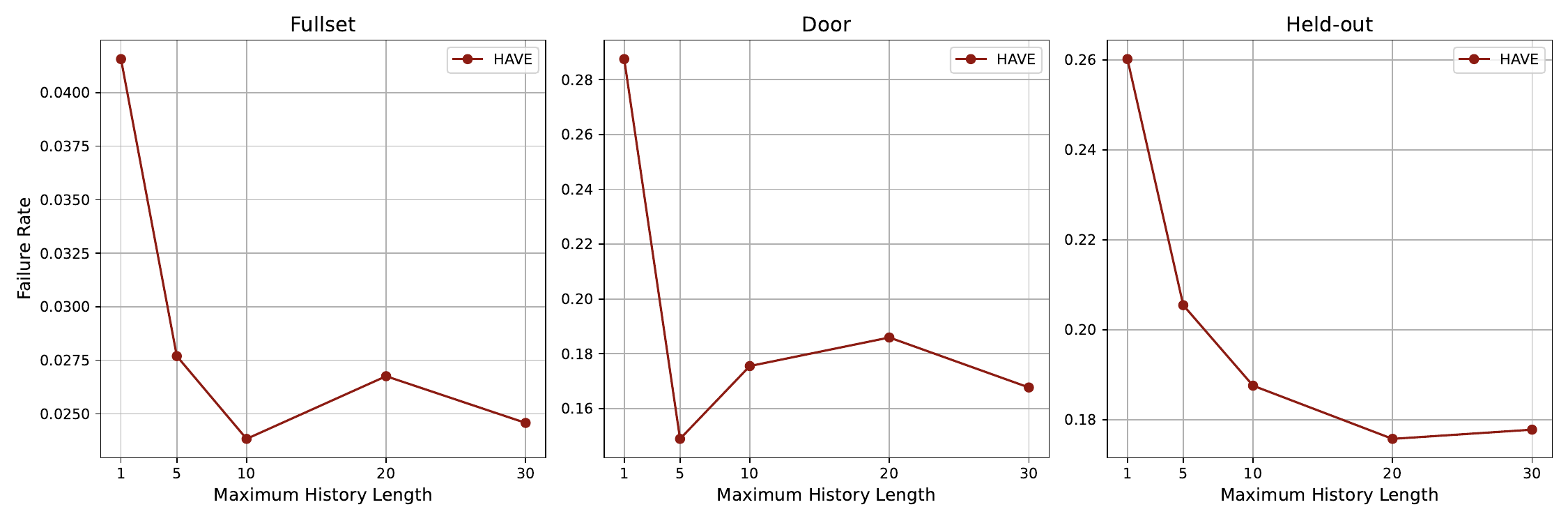} 
    \caption{\textbf{Max History Length Analysis}: Performance improves with longer history up to a point, but excessive context can introduce redundancy and instability.}
    \label{fig:max_his_len}
\end{figure}

We experiment in all three articulated objects environments with different maximum history $K$ length during inference time, where we only keep the most recent $K$ step histories. As shown in Fig.~\ref{fig:max_his_len}, we see that the improvement from using 1-step history to 5-step history is significant, further increasing history length improves the performance in fullset and held-out categories, but in a much smaller scale. The intuition behind this is that with a short history length, there might not be enough context for the model to reason with to select the correct mode, but with a long history, the information is often times redundant and harder to process and may induce instability.

\subsection{HAVE with Conditional Generators}

\begin{figure}[h]
    \centering
    \includegraphics[width=0.9\textwidth]{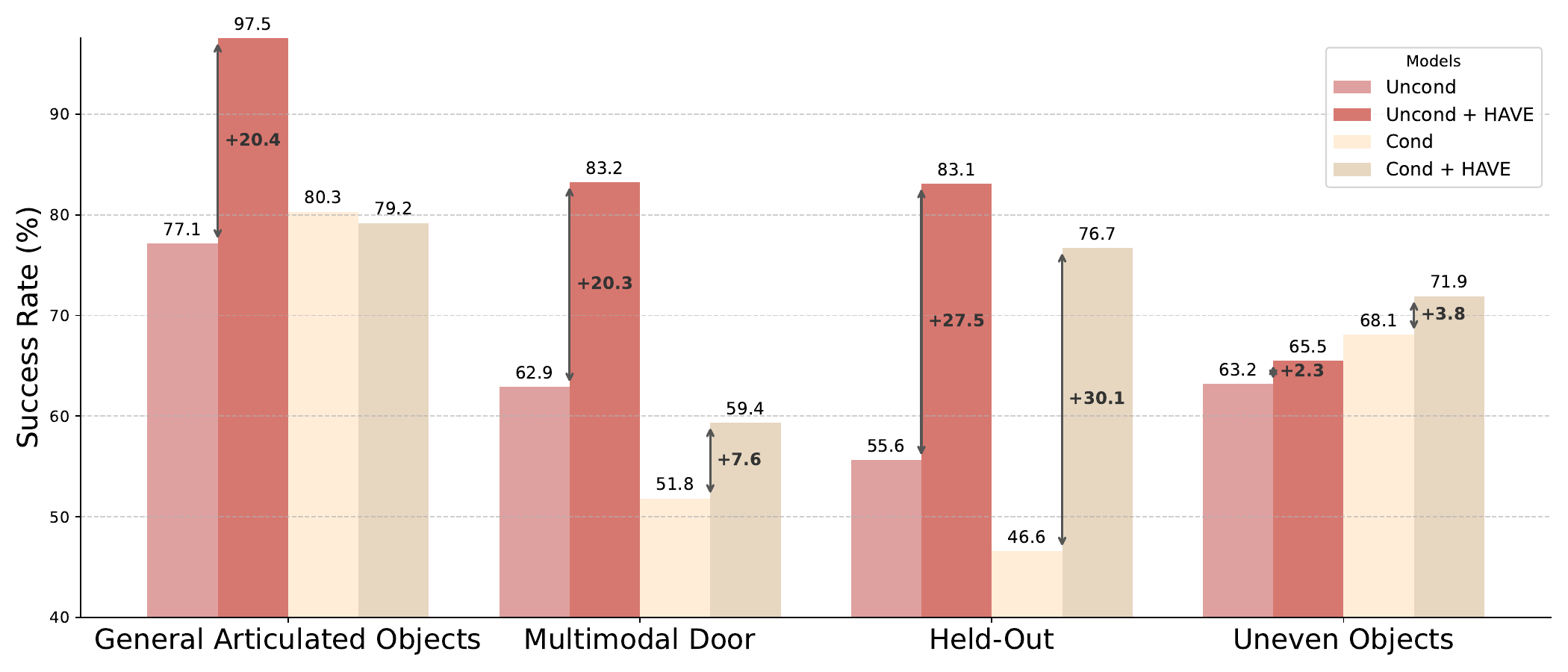} 
    \caption{\textbf{Performance of HAVE with Unconditional and Conditional Generators}: The better expressiveness of the unconditional generator allows HAVE to achieve best performance on complex tasks like 'Articulated Objects', while conditional generator achieves better performance for simpler tasks where the action space is less varied.}
    \label{fig:have_uncond_cond}
\end{figure}

We compared the performance of HAVE when paired with either an unconditional or a conditional diffusion generator. As shown in Fig. \ref{fig:have_uncond_cond}, using an unconditional generator with HAVE consistently yields better results in environments with complex action spaces, such as for articulated objects. A potential reason is that the unconditional generator is more expressive and captures a wider range of actions. In contrast, the conditional generator, which is trained to reduce ambiguity, is more likely to experience mode collapse, limiting the verifier's ability to assist. For environments where the generator is less confident, as with held-out categories, it tends to generate more diverse modes, allowing the verifier to improve performance more significantly compared with other environments. Finally, in the simpler "Uneven Objects" environment, the conditional generator's history-aware structure allows it to model the action distribution more effectively, leading to better performance.

\subsection{Classifier Guidance}

With a diffusion-based generator and a verifier, the most intuitive way is to directly apply the classifier guidance~\cite{dhariwal2021diffusion} to the generation process. Therefore we also experiment with classifier guidance to compare with our proposed method HAVE. Because we generate 3D Articulation Flow~\cite{eisner2022flowbot3d} and verify the dense action field (Sec.5.3), we transform the noisy articulation flow prediction $\{f_i\}$ from each step to a dense action field $\{d_i\}$ to pass through the verifier: We first compute a weighted average of the noisy flow to obtain a global direction $\bar{f}$, and a weighted grasp center $\bar{p}$ from the point cloud, both using flow magnitudes $||f_i||$ as weights. Then we use these point and action direction to create the dense action field following the magnitude decaying equation from Sec.5.3. This process is differentiable and therefore the verifier's gradients over the initial articulation flow generation can be calculated - we then use the gradients for classifier guidance. 

\begin{figure}[h]
    \centering
    \includegraphics[width=1.0\textwidth]{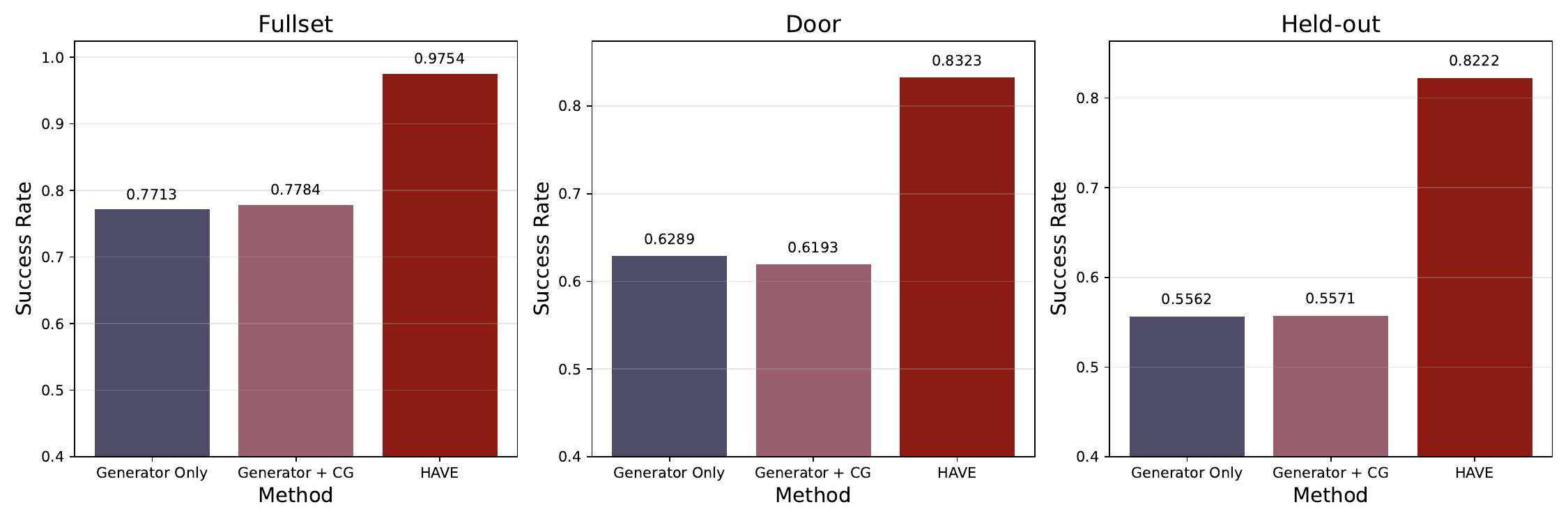} 
    \caption{\textbf{Classifier Guidance Performance}: We plot the \textbf{Success Rate ($\uparrow$)} with and without classifier guidance, and compare them with HAVE.}
    \label{fig:classifier_guidance}
\end{figure}

We can see from Fig.~\ref{fig:classifier_guidance} the classifier guidance offers a minor performance improvement, and has a big gap with HAVE. One possible reason is that even though classifier guidance provides gradients over intermediate states, these gradients can be misleading: the transformation from noisy flow ${f_i}$ to a clean dense action field ${d_i}$ may distort the gradient signal, and even without this transformation, the intermediate noisy flows themselves are out-of-distribution for the verifier, which is trained only on clean data. In contrast, HAVE naturally reasons over history trajectories and scores action proposals, offering a more reliable and effective approach for selecting good actions.

\subsection{CFG scale for Conditional Diffusion}
\label{sec:cfg_for_cond_diff}

\begin{figure}[h]
    \centering
    \includegraphics[width=1.0\textwidth]{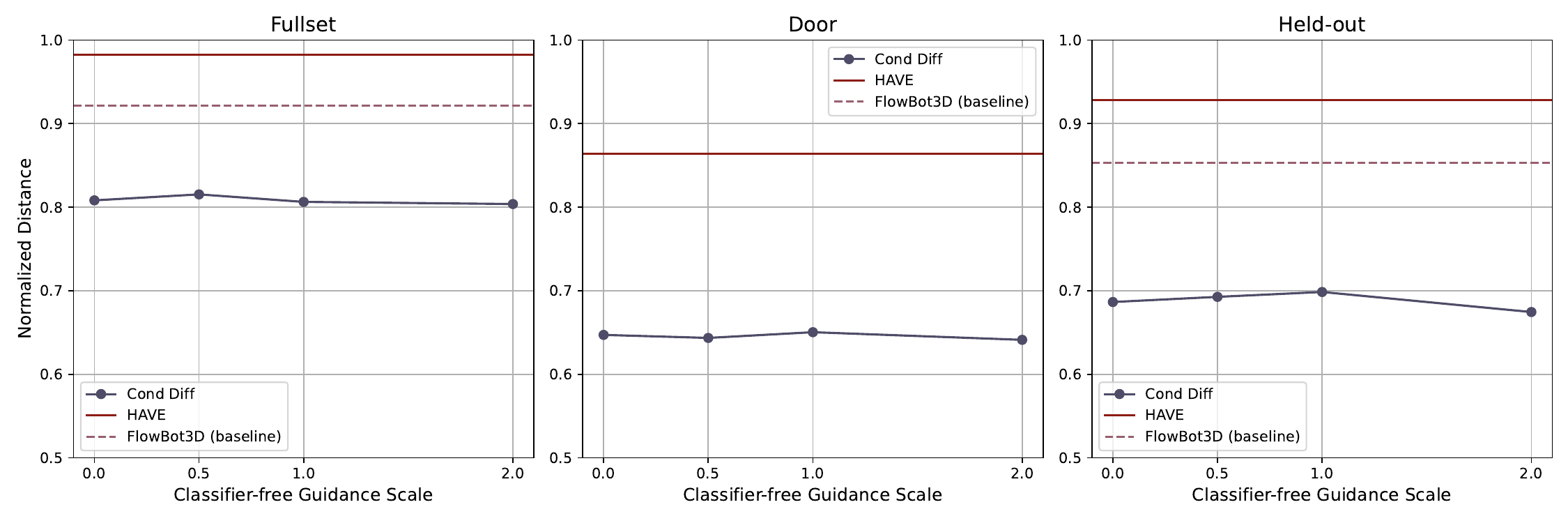} 
    \caption{\textbf{Classifier-Free Guidance Scale Analysis}: We plot the \textbf{Normalized Distance ($\uparrow$)} against different classifier-free guidance scale, and the performance of HAVE and FlowBot3D as reference.}
    \label{fig:cond_diffusion_cfg}
\end{figure}

We tried different classifier-free guidance scale at test time for the conditional diffusion baseline as shown in Fig.~\ref{fig:cond_diffusion_cfg}. In the figure, we plot \textbf{Normalized Distance} (the normalized distance to being fully open, as used in ~\citet{eisner2022flowbot3d}) against different classifier-free guidance scale. From the curves, we see that the best performance is achieved with classifier-free guidance scale of 0.5 or 1.0 (varies across tasks), but the overall performance is about the same. This indicates that the current conditional diffusion structure is not understanding the histories well, demonstrating that training a history-aware conditional diffusion model is practically harder and less efficient than training a verifier.

\subsection{Generalization with Different Verifier Capacity}

We test the generalization ability of the verifier and analyze whether overfitting occurs by comparing training and test metrics. As shown in Table~\ref{tab:reb_overfit}, the verifier demonstrates strong generalization to unseen instances and reasonable generalization to unseen categories, though performance drops more in the latter case—where overfitting is most noticable. 

\begin{table}[h]
    \centering
    \caption{Overfitting Analysis (Failure Rate \%; Averaged across class)}
    \resizebox{0.5\linewidth}{!}{
        \begin{tabular}{lcccc}
            \toprule
             & General & Doors & Held-out & Uneven \\
            \midrule
            Train set & 1.84 & 2.96 & 3.69 & 12.20 \\ 
            Test set & 2.26 & 16.77 & 16.92 & 12.93 \\ 
            \bottomrule
        \end{tabular}
    }
    \label{tab:reb_overfit}
\end{table}

To investigate how model capacity contributes to this, we varied the transformer size and compared the failure rates on train and unseen categories in Table~\ref{tab:reb_capacity}. We use 4 attention heads, 4 attention layers and 256 as feature dimension for the final model. We found that a larger model (8 attention heads, 6 attention layers, and feature dimension as 512) slightly improves training accuracy but hurts generalization, suggesting it may be too large for the task and leads to overfitting. A smaller model (2 attention heads, 2 attention layers, and feature dimension as 256) performs comparably on unseen categories, suggesting that model capacity has limited impact and our chosen size is appropriate for the task.

\begin{table}[h]
    \centering
    \caption{Capacity Analysis on Held-Out (Failure Rate \%); Metric presented as $AVG_{\text{sample}}$($AVG_{\text{class}}$)}
    \resizebox{0.7\linewidth}{!}{
        \begin{tabular}{lccc}
            \toprule
             & Original & Smaller & Bigger  \\
            \midrule
            Train categories & 2.49 (3.69) & 2.86 (3.37) & 2.16 (2.54) \\
            Unseen categories & 11.31 (16.92) & 11.03 (18.58) & 12.70 (20.57) \\
            \bottomrule
        \end{tabular}
    }
    \label{tab:reb_capacity}
\end{table}

%% file: tables/fr_train-val-unseen_unseen.tex
\begin{table*}[ht]
\newcommand{\clipart}[1]{\includegraphics[height=6mm]{tables/icons/#1.png}}
\renewcommand{\arraystretch}{1.2}
\resizebox{\textwidth}{!}{
\setlength\tabcolsep{.2em}

\begin{tabular}{|r|c|c|cccccccccc|}
\hline
\rule{0pt}{2.5em}
& \textbf{\underline{$\textbf{AVG}_{c}$}} & \textbf{\underline{$\textbf{AVG}_{s}$}} 
& \clipart{box} & \clipart{phone} & \clipart{dish} & \clipart{safe} & \clipart{oven} & \clipart{washer} & \clipart{table} & \clipart{pot} & \clipart{bucket} & \clipart{door} \\
\hline
\textbf{Baselines} & \multicolumn{12}{c|}{} \\
\hline
FlowBot3D & 39.6 & 29.4 & 56.0 & 66.7 & 38.6 & 27.9 & 28.5 & \textbf{14.1} & 25.0 & \textbf{0.0} & 96.7 & 42.4 \\
FlowBotHD (w/o CC) & 30.4 & 28.1 & 46.7 & 6.7 & 13.8 & 37.9 & 6.7 & 45.9 & 31.0 & \textbf{0.0} & 73.3 & 42.4 \\
FlowBotHD (w/ CC) & \textbf{14.8} & 10.9 & \textbf{22.7} & \textbf{0.0} & 6.2 & 32.9 & \textbf{3.6} & 15.3 & 4.6 & \textbf{0.0} & 30.0 & \textbf{32.9} \\
Generator Only & 44.4 & 25.0 & 85.3 & 66.7 & 14.8 & 65.7 & 30.9 & 44.7 & 8.0 & 0.8 & 83.3 & 43.5 \\
Conditional Diffusion & 53.4 & 41.2 & 78.7 & 66.7 & 46.7 & 60.7 & 58.8 & 35.3 & 26.6 & \textbf{0.0} & 86.7 & 74.1\\
\hline
\textbf{Ours} & \multicolumn{12}{c|}{} \\
\hline
HAVE (Ours) + GT obs flow & 16.9 & 11.3 & 44.0 & 13.3 & \textbf{0.5} & 13.6 & 17.6 & 27.1 & 3.2 & \textbf{0.0} & 10.0 & 40.0 \\
HAVE (Ours) + DELTA & 21.5 & 14.1 & 62.7 & 6.7 & 2.4 & 20.0 & 20.6 & 35.3 & 4.2 & \textbf{0.0} & 23.3 & 40.0 \\
\hline
\textbf{Ours (Oracle)} & \multicolumn{12}{c|}{} \\
\hline
Oracle Sampler & 12.2 & 9.5 & 30.7 & 0.0 & 2.4 & 10.7 & 11.5 & 17.7 & 2.4 & 0.0 & 3.3 & 42.9 \\
Oracle Verifier & 14.4 & 9.2 & 58.7 & 6.7 & 1.0 & 10.7 & 6.1 & 23.5 & 2.4 & 0.0 & 3.3 & 31.2 \\
\hline
\textbf{Ablations} & \multicolumn{12}{c|}{} \\
\hline
w/o Unconditional Score Loss & 16.1 & \textbf{10.6} & 60.0 & 6.7 & 1.0 & 14.3 & 7.9 & 27.1 & 2.6 & \textbf{0.0} & \textbf{3.3} & 38.2\\
w/o History Score Loss & 18.1 & \textbf{10.6} & 52.0 & \textbf{0.0} & \textbf{0.5} & \textbf{12.1} & 9.1 & 29.4 & \textbf{2.2} & \textbf{0.0} & 36.7 & 38.8\\
Point Cloud as Result & 23.5 & 13.0 & 62.7 & 6.7 & 3.8 & 19.3 & 12.1 & 33.0 & 2.6 & \textbf{0.0} & 56.7 & 38.2\\
Sparse Action & 22.7 & 12.5 & 72.0 & 33.3 & 1.9 & 15.0 & 9.1 & 28.2 & 3.1 & 0.8 & 23.3 & 40.6 \\
\hline
\end{tabular}
}
\smallskip
\caption{\textbf{Failure Rate \% ($\downarrow$) on Held-out Categories}: Lower is better. AVG$_c$ is the category-wise average, AVG$_s$ is sample-wise average. HAVE provides substantial improvements over baselines without a verifier.}
\label{tab:tvu-failure-test}
\vspace*{-10pt}
\end{table*}

%% file: tables/mimicgen.tex
\begin{table*}[ht]
\centering
\begin{tabular}{r|ccc}
\toprule[1pt]
High-level Type & Square D2 & Nut Assembly D0 & Threading D2\\
\midrule[0.5pt]
TAX3D  &  0.29$\pm$0.03 & 0.11$\pm$0.08 & 0.10$\pm$0.03\\
TAX3D+Verifier  &  \textbf{0.41$\pm$0.12} & \textbf{0.15$\pm$0.13} & \textbf{0.19$\pm$0.07}\\
\bottomrule[1pt]
\end{tabular}

\caption{\textbf{Success Rate on MimicGen}~\cite{mandlekar2023mimicgen}: benchmark tasks ($\uparrow$). We train on 3 seeds for each task.}
\label{Table:mimicgen}
\end{table*}

%% file: main.bbl
\begin{thebibliography}{41}
\providecommand{\natexlab}[1]{#1}
\providecommand{\url}[1]{\texttt{#1}}
\expandafter\ifx\csname urlstyle\endcsname\relax
  \providecommand{\doi}[1]{doi: #1}\else
  \providecommand{\doi}{doi: \begingroup \urlstyle{rm}\Url}\fi

\bibitem[Setlur et~al.(2025)Setlur, Rajaraman, Levine, and
  Kumar]{setlur2025scaling}
A.~Setlur, N.~Rajaraman, S.~Levine, and A.~Kumar.
\newblock Scaling test-time compute without verification or rl is suboptimal.
\newblock \emph{arXiv preprint arXiv:2502.12118}, 2025.

\bibitem[Bohg et~al.(2017)Bohg, Hausman, Sankaran, Brock, Kragic, Schaal, and
  Sukhatme]{bohg2017interactive}
J.~Bohg, K.~Hausman, B.~Sankaran, O.~Brock, D.~Kragic, S.~Schaal, and G.~S.
  Sukhatme.
\newblock Interactive perception: Leveraging action in perception and
  perception in action.
\newblock \emph{IEEE Transactions on Robotics}, 33\penalty0 (6):\penalty0
  1273--1291, 2017.

\bibitem[Huang et~al.(2023)Huang, Zhang, Cao, Liu, Xu, Ding, Francis, Chen, and
  Zhao]{huang2023went}
P.~Huang, X.~Zhang, Z.~Cao, S.~Liu, M.~Xu, W.~Ding, J.~Francis, B.~Chen, and
  D.~Zhao.
\newblock What went wrong? closing the sim-to-real gap via differentiable
  causal discovery.
\newblock In \emph{Conference on Robot Learning}, pages 734--760. PMLR, 2023.

\bibitem[Katz and Brock(2008)]{katz2008articulate}
D.~Katz and O.~Brock.
\newblock Manipulating articulated objects with interactive perception.
\newblock In \emph{2008 IEEE International Conference on Robotics and
  Automation}, pages 272--277, 2008.
\newblock \doi{10.1109/ROBOT.2008.4543220}.

\bibitem[Weng et~al.(2024)Weng, Zhou, Yin, Kravberg, Varava, Navarro-Alarcon,
  and Kragic]{weng2024interactive}
Z.~Weng, P.~Zhou, H.~Yin, A.~Kravberg, A.~Varava, D.~Navarro-Alarcon, and
  D.~Kragic.
\newblock Interactive perception for deformable object manipulation.
\newblock \emph{IEEE Robotics and Automation Letters}, 2024.

\bibitem[Allevato et~al.(2020)Allevato, Short, Pryor, and
  Thomaz]{allevato2020tunenet}
A.~Allevato, E.~S. Short, M.~Pryor, and A.~Thomaz.
\newblock Tunenet: One-shot residual tuning for system identification and
  sim-to-real robot task transfer.
\newblock In \emph{Conference on Robot Learning}, pages 445--455. PMLR, 2020.

\bibitem[Xu et~al.(2022)Xu, Shen, Zhang, Lu, Zhao, Tenenbaum, and
  Gan]{xu2022prompting}
M.~Xu, Y.~Shen, S.~Zhang, Y.~Lu, D.~Zhao, J.~Tenenbaum, and C.~Gan.
\newblock Prompting decision transformer for few-shot policy generalization.
\newblock In \emph{international conference on machine learning}, pages
  24631--24645. PMLR, 2022.

\bibitem[Laskin et~al.(2022)Laskin, Wang, Oh, Parisotto, Spencer, Steigerwald,
  Strouse, Hansen, Filos, Brooks, et~al.]{laskin2022context}
M.~Laskin, L.~Wang, J.~Oh, E.~Parisotto, S.~Spencer, R.~Steigerwald,
  D.~Strouse, S.~Hansen, A.~Filos, E.~Brooks, et~al.
\newblock In-context reinforcement learning with algorithm distillation.
\newblock \emph{arXiv preprint arXiv:2210.14215}, 2022.

\bibitem[Wang et~al.(2025)Wang, Zhang, Wu, Li, Shen, Wu, He, Wang, and
  Dong]{wang2025adamanip}
Y.~Wang, X.~Zhang, R.~Wu, Y.~Li, Y.~Shen, M.~Wu, Z.~He, Y.~Wang, and H.~Dong.
\newblock Adamanip: Adaptive articulated object manipulation environments and
  policy learning.
\newblock \emph{arXiv preprint arXiv:2502.11124}, 2025.

\bibitem[Zhang et~al.(2025)Zhang, Liu, Huang, Han, Lyu, Xu, and
  Zhao]{zhang2025dynamics}
X.~Zhang, S.~Liu, P.~Huang, W.~J. Han, Y.~Lyu, M.~Xu, and D.~Zhao.
\newblock Dynamics as prompts: In-context learning for sim-to-real system
  identifications.
\newblock \emph{IEEE Robotics and Automation Letters}, 2025.

\bibitem[Goodfellow et~al.(2020)Goodfellow, Pouget-Abadie, Mirza, Xu,
  Warde-Farley, Ozair, Courville, and Bengio]{goodfellow2020generative}
I.~Goodfellow, J.~Pouget-Abadie, M.~Mirza, B.~Xu, D.~Warde-Farley, S.~Ozair,
  A.~Courville, and Y.~Bengio.
\newblock Generative adversarial networks.
\newblock \emph{Communications of the ACM}, 63\penalty0 (11):\penalty0
  139--144, 2020.

\bibitem[Christiano et~al.(2017)Christiano, Leike, Brown, Martic, Legg, and
  Amodei]{christiano2017deep}
P.~F. Christiano, J.~Leike, T.~Brown, M.~Martic, S.~Legg, and D.~Amodei.
\newblock Deep reinforcement learning from human preferences.
\newblock \emph{Advances in neural information processing systems}, 30, 2017.

\bibitem[Bai et~al.(2022)Bai, Jones, Ndousse, Askell, Chen, DasSarma, Drain,
  Fort, Ganguli, Henighan, et~al.]{bai2022training}
Y.~Bai, A.~Jones, K.~Ndousse, A.~Askell, A.~Chen, N.~DasSarma, D.~Drain,
  S.~Fort, D.~Ganguli, T.~Henighan, et~al.
\newblock Training a helpful and harmless assistant with reinforcement learning
  from human feedback.
\newblock \emph{arXiv preprint arXiv:2204.05862}, 2022.

\bibitem[Ouyang et~al.(2022)Ouyang, Wu, Jiang, Almeida, Wainwright, Mishkin,
  Zhang, Agarwal, Slama, Ray, et~al.]{ouyang2022training}
L.~Ouyang, J.~Wu, X.~Jiang, D.~Almeida, C.~Wainwright, P.~Mishkin, C.~Zhang,
  S.~Agarwal, K.~Slama, A.~Ray, et~al.
\newblock Training language models to follow instructions with human feedback.
\newblock \emph{Advances in neural information processing systems},
  35:\penalty0 27730--27744, 2022.

\bibitem[Swamy et~al.(2025)Swamy, Choudhury, Sun, Wu, and
  Bagnell]{swamy2025all}
G.~Swamy, S.~Choudhury, W.~Sun, Z.~S. Wu, and J.~A. Bagnell.
\newblock All roads lead to likelihood: The value of reinforcement learning in
  fine-tuning.
\newblock \emph{arXiv preprint arXiv:2503.01067}, 2025.

\bibitem[Cobbe et~al.(2021)Cobbe, Kosaraju, Bavarian, Chen, Jun, Kaiser,
  Plappert, Tworek, Hilton, Nakano, et~al.]{cobbe2021training}
K.~Cobbe, V.~Kosaraju, M.~Bavarian, M.~Chen, H.~Jun, L.~Kaiser, M.~Plappert,
  J.~Tworek, J.~Hilton, R.~Nakano, et~al.
\newblock Training verifiers to solve math word problems.
\newblock \emph{arXiv preprint arXiv:2110.14168}, 2021.

\bibitem[Setlur et~al.(2024)Setlur, Nagpal, Fisch, Geng, Eisenstein, Agarwal,
  Agarwal, Berant, and Kumar]{setlur2024rewarding}
A.~Setlur, C.~Nagpal, A.~Fisch, X.~Geng, J.~Eisenstein, R.~Agarwal, A.~Agarwal,
  J.~Berant, and A.~Kumar.
\newblock Rewarding progress: Scaling automated process verifiers for llm
  reasoning.
\newblock \emph{arXiv preprint arXiv:2410.08146}, 2024.

\bibitem[Hosseini et~al.(2024)Hosseini, Yuan, Malkin, Courville, Sordoni, and
  Agarwal]{hosseini2024v}
A.~Hosseini, X.~Yuan, N.~Malkin, A.~Courville, A.~Sordoni, and R.~Agarwal.
\newblock V-star: Training verifiers for self-taught reasoners.
\newblock \emph{arXiv preprint arXiv:2402.06457}, 2024.

\bibitem[Rafailov et~al.(2023)Rafailov, Sharma, Mitchell, Manning, Ermon, and
  Finn]{rafailov2023direct}
R.~Rafailov, A.~Sharma, E.~Mitchell, C.~D. Manning, S.~Ermon, and C.~Finn.
\newblock Direct preference optimization: Your language model is secretly a
  reward model.
\newblock \emph{Advances in Neural Information Processing Systems},
  36:\penalty0 53728--53741, 2023.

\bibitem[Singh et~al.(2023)Singh, Co-Reyes, Agarwal, Anand, Patil, Garcia, Liu,
  Harrison, Lee, Xu, et~al.]{singh2023beyond}
A.~Singh, J.~D. Co-Reyes, R.~Agarwal, A.~Anand, P.~Patil, X.~Garcia, P.~J. Liu,
  J.~Harrison, J.~Lee, K.~Xu, et~al.
\newblock Beyond human data: Scaling self-training for problem-solving with
  language models.
\newblock \emph{arXiv preprint arXiv:2312.06585}, 2023.

\bibitem[Song et~al.(2024)Song, Zhang, Eisenach, Kakade, Foster, and
  Ghai]{song2024mind}
Y.~Song, H.~Zhang, C.~Eisenach, S.~Kakade, D.~Foster, and U.~Ghai.
\newblock Mind the gap: Examining the self-improvement capabilities of large
  language models.
\newblock \emph{arXiv preprint arXiv:2412.02674}, 2024.

\bibitem[Ames et~al.(2019)Ames, Coogan, Egerstedt, Notomista, Sreenath, and
  Tabuada]{ames2019control}
A.~D. Ames, S.~Coogan, M.~Egerstedt, G.~Notomista, K.~Sreenath, and P.~Tabuada.
\newblock Control barrier functions: Theory and applications.
\newblock In \emph{2019 18th European control conference (ECC)}, pages
  3420--3431. Ieee, 2019.

\bibitem[Wang et~al.(2024)Wang, Wang, Du, Sundaralingam, Yang, Chao,
  Perez-D'Arpino, Fox, and Shah]{wang2024inference}
Y.~Wang, L.~Wang, Y.~Du, B.~Sundaralingam, X.~Yang, Y.-W. Chao,
  C.~Perez-D'Arpino, D.~Fox, and J.~Shah.
\newblock Inference-time policy steering through human interactions.
\newblock \emph{arXiv preprint arXiv:2411.16627}, 2024.

\bibitem[Jeong et~al.(2024)Jeong, Chen, and Bajcsy]{jeong2024robots}
H.~J. Jeong, R.~Chen, and A.~Bajcsy.
\newblock Robots that suggest safe alternatives.
\newblock \emph{arXiv preprint arXiv:2409.09883}, 2024.

\bibitem[Nakamura et~al.(2025)Nakamura, Peters, and
  Bajcsy]{nakamura2025generalizing}
K.~Nakamura, L.~Peters, and A.~Bajcsy.
\newblock Generalizing safety beyond collision-avoidance via latent-space
  reachability analysis.
\newblock \emph{arXiv preprint arXiv:2502.00935}, 2025.

\bibitem[Wu et~al.(2025)Wu, Tian, Swamy, and Bajcsy]{wu2025foresight}
Y.~Wu, R.~Tian, G.~Swamy, and A.~Bajcsy.
\newblock From foresight to forethought: Vlm-in-the-loop policy steering via
  latent alignment.
\newblock \emph{arXiv preprint arXiv:2502.01828}, 2025.

\bibitem[Borquez et~al.(2025)Borquez, Raus, Ciftci, and
  Bansal]{borquez2025dualguard}
J.~Borquez, L.~Raus, Y.~U. Ciftci, and S.~Bansal.
\newblock Dualguard mppi: Safe and performant optimal control by combining
  sampling-based mpc and hamilton-jacobi reachability.
\newblock \emph{arXiv preprint arXiv:2502.01924}, 2025.

\bibitem[Qi et~al.(2017)Qi, Yi, Su, and Guibas]{qi2017pn2}
C.~R. Qi, L.~Yi, H.~Su, and L.~J. Guibas.
\newblock Pointnet++: Deep hierarchical feature learning on point sets in a
  metric space.
\newblock In I.~Guyon, U.~V. Luxburg, S.~Bengio, H.~Wallach, R.~Fergus,
  S.~Vishwanathan, and R.~Garnett, editors, \emph{Advances in Neural
  Information Processing Systems}, volume~30. Curran Associates, Inc., 2017.

\bibitem[Ngo et~al.(2024)Ngo, Zhuang, Gan, Kalogerakis, Tulyakov, Lee, and
  Wang]{ngo2024delta}
T.~D. Ngo, P.~Zhuang, C.~Gan, E.~Kalogerakis, S.~Tulyakov, H.-Y. Lee, and
  C.~Wang.
\newblock Delta: Dense efficient long-range 3d tracking for any video.
\newblock \emph{arXiv preprint arXiv:2410.24211}, 2024.

\bibitem[Li et~al.(2024)Li, Leng, Fang, Eisner, and Held]{li2024flowbothd}
Y.~Li, W.~H. Leng, Y.~Fang, B.~Eisner, and D.~Held.
\newblock Flowbot{HD}: History-aware diffuser handling ambiguities in
  articulated objects manipulation.
\newblock In \emph{8th Annual Conference on Robot Learning}, 2024.

\bibitem[Eisner et~al.(2022)Eisner, Zhang, and Held]{eisner2022flowbot3d}
B.~Eisner, H.~Zhang, and D.~Held.
\newblock Flowbot3d: Learning 3d articulation flow to manipulate articulated
  objects.
\newblock \emph{arXiv preprint arXiv:2205.04382}, 2022.

\bibitem[Ho and Salimans(2022)]{ho2022classifier}
J.~Ho and T.~Salimans.
\newblock Classifier-free diffusion guidance.
\newblock \emph{arXiv preprint arXiv:2207.12598}, 2022.

\bibitem[Leadbetter et~al.(2012)Leadbetter, Lindgren, and
  Rootz{\'e}n]{leadbetter2012extremes}
M.~R. Leadbetter, G.~Lindgren, and H.~Rootz{\'e}n.
\newblock \emph{Extremes and related properties of random sequences and
  processes}.
\newblock Springer Science \& Business Media, 2012.

\bibitem[DasGupta et~al.(2014)DasGupta, Lahiri, and
  Stoyanov]{dasgupta2014sharp}
A.~DasGupta, S.~Lahiri, and J.~Stoyanov.
\newblock Sharp fixed n bounds and asymptotic expansions for the mean and the
  median of a gaussian sample maximum, and applications to the donoho--jin
  model.
\newblock \emph{Statistical Methodology}, 20:\penalty0 40--62, 2014.

\bibitem[Peebles and Xie(2023)]{peebles2023scalable}
W.~Peebles and S.~Xie.
\newblock Scalable diffusion models with transformers.
\newblock In \emph{Proceedings of the IEEE/CVF International Conference on
  Computer Vision}, pages 4195--4205, 2023.

\bibitem[Xiang et~al.(2020)Xiang, Qin, Mo, Xia, Zhu, Liu, Liu, Jiang, Yuan,
  Wang, et~al.]{xiang2020sapien}
F.~Xiang, Y.~Qin, K.~Mo, Y.~Xia, H.~Zhu, F.~Liu, M.~Liu, H.~Jiang, Y.~Yuan,
  H.~Wang, et~al.
\newblock Sapien: A simulated part-based interactive environment.
\newblock In \emph{Proceedings of the IEEE/CVF conference on computer vision
  and pattern recognition}, pages 11097--11107, 2020.

\bibitem[Mandlekar et~al.(2023)Mandlekar, Nasiriany, Wen, Akinola, Narang, Fan,
  Zhu, and Fox]{mandlekar2023mimicgen}
A.~Mandlekar, S.~Nasiriany, B.~Wen, I.~Akinola, Y.~Narang, L.~Fan, Y.~Zhu, and
  D.~Fox.
\newblock Mimicgen: A data generation system for scalable robot learning using
  human demonstrations.
\newblock In \emph{7th Annual Conference on Robot Learning}, 2023.

\bibitem[Cai et~al.(2024)Cai, Donca, Eisner, and Held]{cai2024tax3d}
E.~Cai, O.~Donca, B.~Eisner, and D.~Held.
\newblock Non-rigid relative placement through 3d dense diffusion.
\newblock In \emph{8th Annual Conference on Robot Learning}, 2024.
\newblock URL \url{https://arxiv.org/abs/2410.19247}.

\bibitem[Ze et~al.(2024)Ze, Zhang, Zhang, Hu, Wang, and Xu]{Ze2024DP3}
Y.~Ze, G.~Zhang, K.~Zhang, C.~Hu, M.~Wang, and H.~Xu.
\newblock 3d diffusion policy: Generalizable visuomotor policy learning via
  simple 3d representations.
\newblock In \emph{Proceedings of Robotics: Science and Systems (RSS)}, 2024.

\bibitem[Su et~al.(2024)Su, Ahmed, Lu, Pan, Bo, and Liu]{su2024roformer}
J.~Su, M.~Ahmed, Y.~Lu, S.~Pan, W.~Bo, and Y.~Liu.
\newblock Roformer: Enhanced transformer with rotary position embedding.
\newblock \emph{Neurocomputing}, 568:\penalty0 127063, 2024.

\bibitem[Dhariwal and Nichol(2021)]{dhariwal2021diffusion}
P.~Dhariwal and A.~Nichol.
\newblock Diffusion models beat gans on image synthesis.
\newblock \emph{Advances in neural information processing systems},
  34:\penalty0 8780--8794, 2021.

\end{thebibliography}
